\pdfoutput=1

\documentclass[11pt,twocolumn,letterpaper]{article}
\PassOptionsToPackage{table}{xcolor}

\usepackage[final]{acl}

\usepackage{times}
\usepackage{latexsym}

\usepackage[T1]{fontenc}

\usepackage[utf8]{inputenc}

\usepackage{microtype}

\usepackage{inconsolata}

\usepackage[table]{xcolor}

\usepackage{graphicx}
\usepackage{float}
\usepackage{caption}
\usepackage{lipsum}
\usepackage{booktabs}
\usepackage{multirow}

\newcommand{\blank}[1]{\hspace*{#1}}
\usepackage{algorithm}
\usepackage{algorithmic}
\usepackage{subcaption}

\usepackage[most]{tcolorbox}
\tcbuselibrary{skins,breakable} 

\definecolor{bittersweet}{rgb}{1.0, 0.44, 0.37}
\definecolor{mygreen}{rgb}{0.29, 0.7, 0.48}
\usepackage{pifont}
\definecolor{demphcolor}{RGB}{144,144,144}

\definecolor{mygray}{gray}{0.4}
\definecolor{autopurple}{HTML}{7030A0}
\definecolor{dyna_yellow}{HTML}{BF9000}
\definecolor{adaptive_blue}{HTML}{0070C0}
\definecolor{darksalmon}{rgb}{0.91, 0.59, 0.48}
\definecolor{emerald}{rgb}{0.31, 0.78, 0.47}
\definecolor{green(pigment)}{rgb}{0.0, 0.65, 0.31}
\definecolor{amaranth}{rgb}{0.9, 0.17, 0.31}
\definecolor{iris}{rgb}{0.35, 0.31, 0.81}
\definecolor{uu}{rgb}{0.95, 0.51, 0.51}
\definecolor{spirodiscoball}{rgb}{0.06, 0.75, 0.99}
\definecolor{Periwinkle}{RGB}{204,204,255} 
\definecolor{mygrey}{RGB}{128,128,128}
\definecolor{TealBlue}{RGB}{0,130,127} 
\definecolor{ForestGreen}{RGB}{34,139,34} 
\definecolor{Melon}{RGB}{253,188,180}        
\definecolor{LightMelon}{RGB}{255,218,206}  

\usepackage{listings}
\definecolor{LightGreen}{RGB}{144,238,144}
\definecolor{jsonkeys}{RGB}{19,110,194}      
\definecolor{jsonstring}{RGB}{42,161,152}    
\definecolor{jsonnumber}{RGB}{211,54,130}    

\lstdefinelanguage{json}{
    string=[s]{"}{"},
    comment=[l]{//},
    morecomment=[s]{/*}{*/},
    keywords={false,true,null},
    sensitive=false
}

\lstdefinestyle{json}{
    language=json,    
    basicstyle=\ttfamily\small,
    numbers=none,
    numberstyle=\small,
    stepnumber=1,
    numbersep=8pt,
    showstringspaces=false,
    breaklines=true,
    frameround=ftff,
    frame=single,
    belowcaptionskip=1\baselineskip,
    keywordstyle=\color{jsonkeys},
    stringstyle=\color{jsonstring},
    numberstyle=\color{jsonnumber},
    identifierstyle=\color{jsonkeys},
    morekeywords={false,true,null}
}

\lstdefinelanguage{JavaScript}{
  keywords={break, case, catch, continue, debugger, default, delete, do, else, false, finally, for, function, if, in, instanceof, new, null, return, switch, this, throw, true, try, typeof, var, void, while, with, let, const},
  morecomment=[l]{//},
  morecomment=[s]{/*}{*/},
  morestring=[b]',
  morestring=[b]",
  sensitive=true
}

\lstdefinestyle{javascript}{
    language=JavaScript,
    backgroundcolor=\color{gray!10},
    basicstyle=\ttfamily,
    keywordstyle=\color{blue},
    stringstyle=\color{red},
    commentstyle=\color{green!60!black},
    numbers=left,
    numberstyle=\tiny\color{gray},
    frame=single,
    breaklines=true,
    breakatwhitespace=true,
    showstringspaces=false
}

\usepackage{listings}
\usepackage{fancyvrb}

\definecolor{ForestGreen}{RGB}{34,139,34}
\definecolor{mygrey}{RGB}{128,128,128}

\lstdefinelanguage{json}{
    basicstyle=\ttfamily\footnotesize,
    numbers=left,
    numberstyle=\tiny,
    stepnumber=1,
    numbersep=8pt,
    showstringspaces=false,
    breaklines=true,
    frame=none,
    backgroundcolor=\color{white},
    literate=
     *{0}{{{\color{blue}0}}}{1}
      {1}{{{\color{blue}1}}}{1}
      {2}{{{\color{blue}2}}}{1}
      {3}{{{\color{blue}3}}}{1}
      {4}{{{\color{blue}4}}}{1}
      {5}{{{\color{blue}5}}}{1}
      {6}{{{\color{blue}6}}}{1}
      {7}{{{\color{blue}7}}}{1}
      {8}{{{\color{blue}8}}}{1}
      {9}{{{\color{blue}9}}}{1}
      {:}{{{\color{red}{:}}}}{1}
      {,}{{{\color{red}{,}}}}{1}
      {\{}{{{\color{red}{\{}}}}{1}
      {\}}{{{\color{red}{\}}}}}{1}
      {[}{{{\color{red}{[}}}}{1}
      {]}{{{\color{red}{]}}}}{1},
}
\title{BannerAgency: Advertising Banner Design with Multimodal LLM Agents}

\author{Heng Wang$^{\dag}$, Yotaro Shimose, Shingo Takamatsu\\\\
Sony Group Corporation \\
\texttt{\{heng.wang, yotaro.shimose, shingo.takamatsu\}@sony.com}
}

\begin{document}

\maketitle

\begin{abstract}
Advertising banners are critical for capturing user attention and enhancing advertising campaign effectiveness. Creating aesthetically pleasing banner designs while conveying the campaign messages is challenging due to the large search space involving multiple design elements. Additionally, advertisers need multiple sizes for different displays and various versions to target different sectors of audiences. Since design is intrinsically an iterative and subjective process, flexible editability is also in high demand for practical usage. While current models have served as assistants to human designers in various design tasks, they typically handle only segments of the creative design process or produce pixel-based outputs that limit editability. This paper introduces a training-free framework for fully automated banner ad design creation, enabling frontier multimodal large language models (MLLMs) to streamline the production of effective banners with minimal manual effort across diverse marketing contexts. We present BannerAgency, an MLLM agent system that collaborates with advertisers to understand their brand identity and banner objectives, generates matching background images, creates blueprints for foreground design elements, and renders the final creatives as editable components in Figma or SVG formats rather than static pixels. To facilitate evaluation and future research, we introduce BannerRequest400, a benchmark featuring 100 unique logos paired with 400 diverse banner requests. Through quantitative and qualitative evaluations, we demonstrate the framework's effectiveness, emphasizing the quality of the generated banner designs, their adaptability to various banner requests, and their strong editability enabled by this component-based approach. 
\end{abstract}

\def\thefootnote{\dag}\footnotetext{Corresponding author: Heng Wang (heng.wang@sony.com, hengwang.vision@gmail.com)} 

\section{Introduction}
\begin{figure}[t]
  \centering
  \begin{subfigure}{\linewidth}
    {\includegraphics[width=\linewidth]{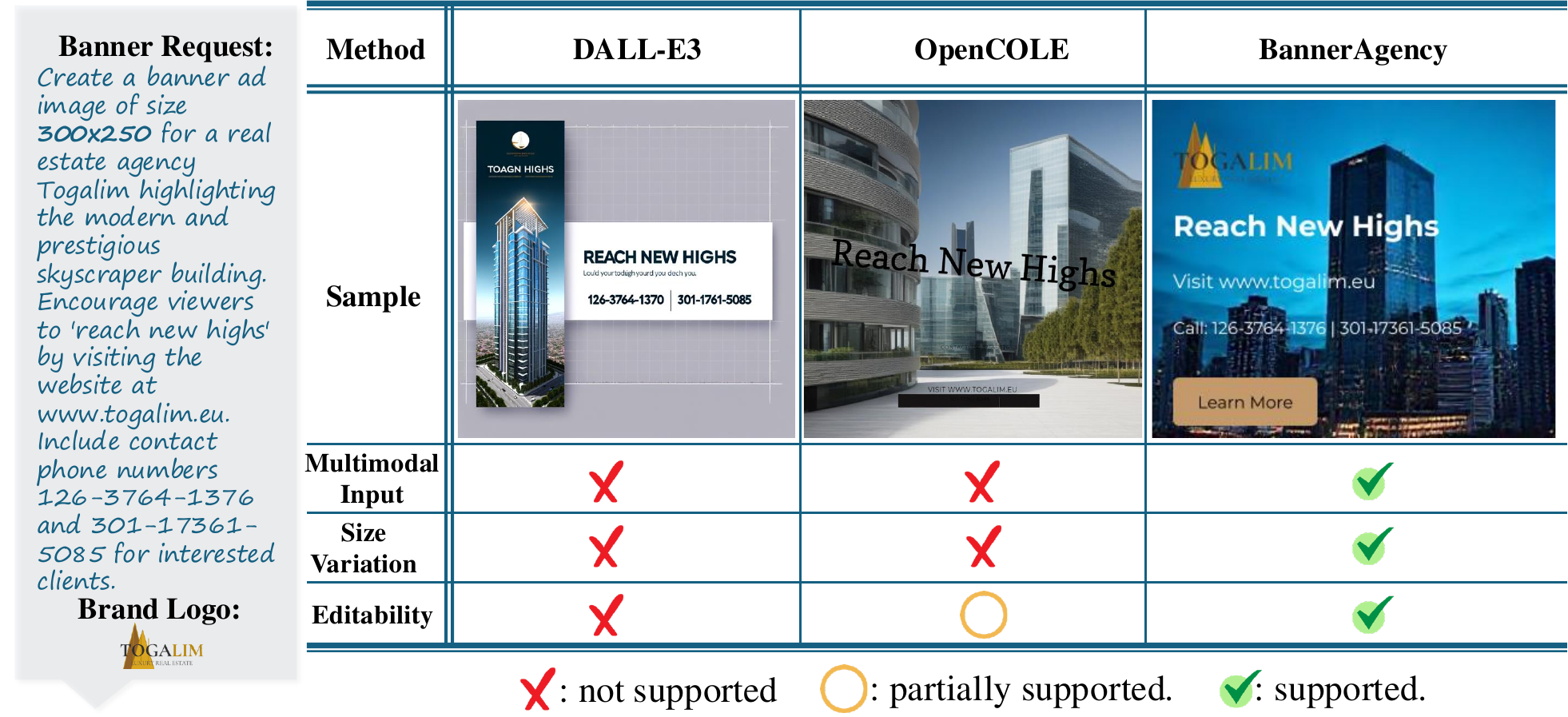}}
    \caption{Left: multimodal banner request. Right: Algorithm comparison.}
    \label{fig:teaserb-a}
  \end{subfigure}
  \hfill
  \begin{subfigure}{\linewidth}
    {\includegraphics[width=\linewidth]{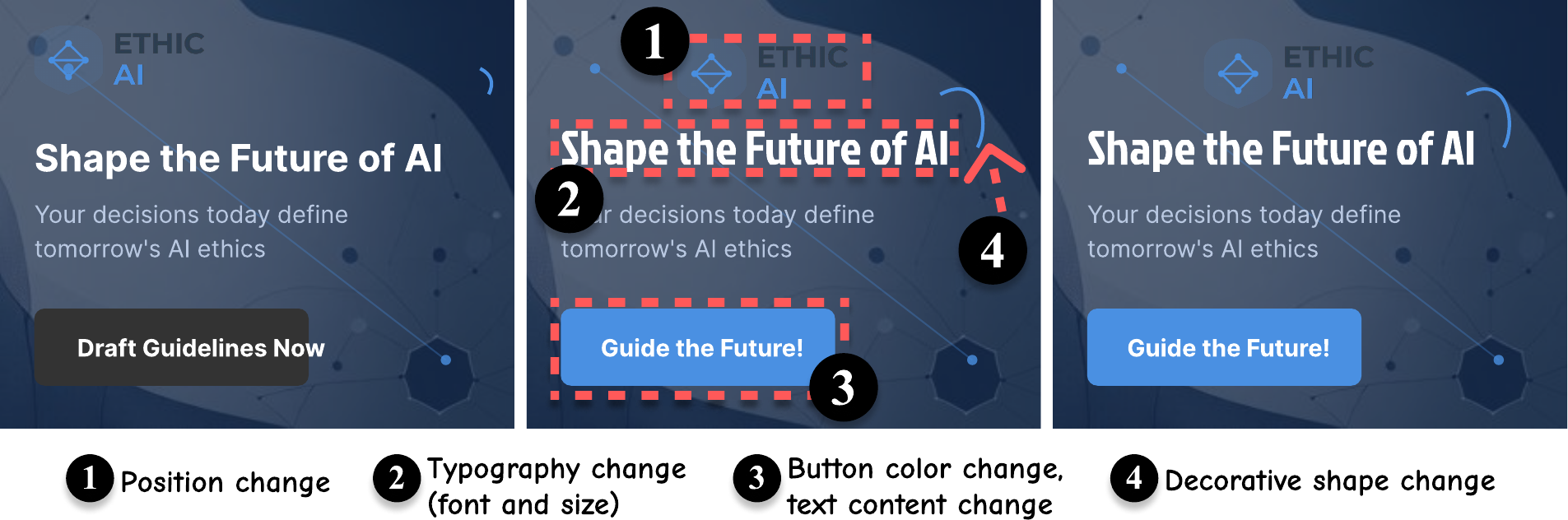}}
    \caption{An example of element-level editing.}
    \label{fig:teaserb-b}
  \end{subfigure}
  \caption{Algorithm comparison and BannerAgency's  editability in banner ad image generation.}
  \label{fig:teaserb}
\end{figure}
Advertising banners constitute an instrumental  medium in digital marketing campaigns. These graphical displays efficiently capture viewer attention, communicate strategic messaging, and enhance brand visibility across competitive market environments~\cite{metricsurvey6,metricsurvey4,chen2021automated,vaddamanu2022harmonized}. Unlike generic images, banner ad images are visually harmonized compositions of multiple creative assets including backdrops, brand-related logos or product images, click-to-action (CTA) buttons, campaign texts, and other decorative elements such as shapes and patterns~\cite{maheshwari2019exemplar,vempati2020enabling}. Campaign texts further require typography design choices, contributing to a vast search space that demands significant time and effort~\cite{du2024towards,creagan}. In addition, advertisers have various size requirements for different display needs~\cite{bannersize} and diverse design requirements tailored to specific target audiences and purposes~\cite{whystrategist}. As design remains inherently creative and subjective, ensuring human interaction and editing capability is crucial for practical applications~\cite{vinci,cole}.

Current models accelerate this banner design process by automating separate aspects:  product background generation~\cite{creagan,yang2024new,fong2024branddiffusion,du2024towards,chen2025ctr}, template-based layout matching~\cite{yang2016automatic,adobeexpress,canvabanner,luban,li2023smartbanner}, layout generation ~\cite{hu2021study,vempati2020enabling,zhang2017layout,maheshwari2019exemplar,lin2024elements}, and automatic typography design~\cite{choi2018fontmatcher,iluz2023word,park2024kinetic}. Advanced scene-text rendering models~\cite{textdiffuser,glyphbyt5,tuoanytext,zhaoharmonizing} and text-to-image (T2I) models such as DALL-E3~\cite{dalle3} are suboptimal for banner design due to their pixel-based generation paradigm, which produces outputs that are difficult to edit and often contains incoherent text as in Fig.~\ref{fig:teaserb-a}. Although TypeR~\cite{typer} addresses the "typos" from T2I models through detection and correction of incorrectly rendered text, it still constrains the design process to a pixel-based approach, limiting subsequent editability. Recent works~\cite{cole,opencole,prompt2poster} explore automatic poster generation from prompts, but banner design requires handling both structured requests and unstructured brand identity like logos. Their data dependency and fine-tuning further limit adaptability to diverse banner sizes and requests. Although these works allow text modification through typography rendering, they merge decorative elements with images, restricting editability (see Fig.\ref{fig:teaserb-a}, where COLE-based\cite{cole,opencole} output's black rectangle cannot be removed).

\begin{figure*}[t]
\centering     
\includegraphics[width=\linewidth]{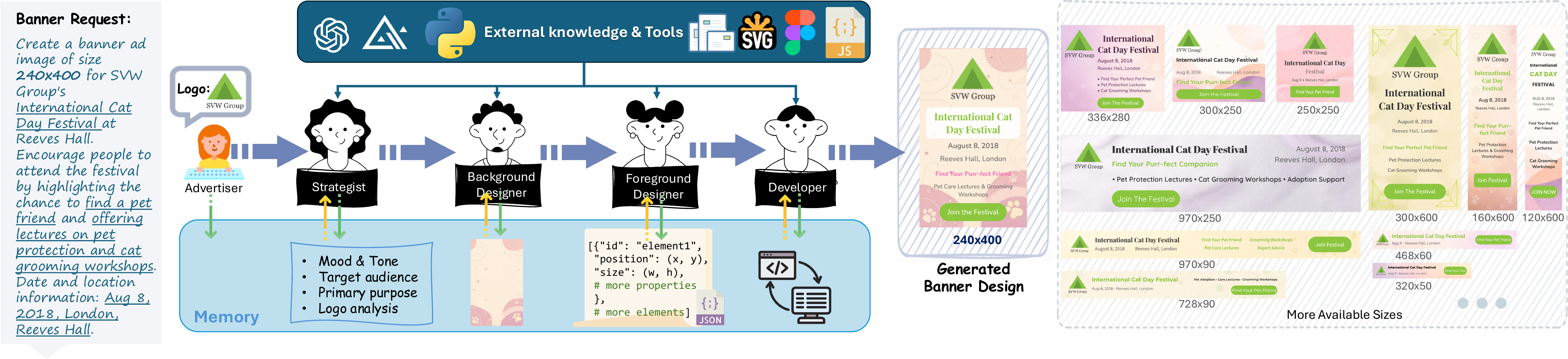}
\caption{\textbf{BannerAgency overview.} Given a logo and request, BannerAgency begins with the \textit{Strategist} analyzing banner objectives, followed by the \textit{Background Designer} creating matching backgrounds, then the \textit{Foreground Designer} producing element blueprints, and concludes with the \textit{Developer} rendering the final design as editable components. With access to external knowledge, tool-calling capabilities, and shared memory, BannerAgency enables context-aware, harmonized decisions and supports multiple banner sizes if requested.}
\label{fig:overview}
\end{figure*}

To bridge these gaps and deliver a fully automatic yet editable banner design solution, we introduce BannerAgency, a training-free framework powered by multimodal large language models (MLLMs)~\cite{gpt4o,claude,mllm1}. Our fundamental idea is to utilize MLLMs as AI agents~\cite{wu2024transagents,zhang2024exploring,islam2024mapcoder,qian2024chatdev} that simulate how a human design team works collaboratively from conception to implementation~\cite{whystrategist}, leveraging the emergent design knowledge inherited from web-scale training data~\cite{layoutprompter,infographic1,genartist}. With tool calling and memory sharing capabilities of LLM agents~\cite{gupta2023visual,lu2024ai,hu2024scenecraft,xu2024magic}, we effectively navigate the extensive design search space through our specialized agent architecture. The system comprises: (1) a \textit{Strategist} that constrains design possibilities by establishing brand guidelines and campaign objectives; (2) a \textit{Background Designer} responsible for visual backdrop generation; (3) a \textit{Foreground Designer} that creates blueprints for banner assets with engaging content and harmonized styling and positioning; and (4) a \textit{Developer} that transforms these specifications into editable components via Figma plugin code or SVG generation. Unlike pixel-based methods that produce rastic outputs, BannerAgency generates designs directly within professional tools, enabling seamless editing of major design components---text, typography, buttons, shapes, and layouts---within a familiar interface used by professional designers as illustrated in Fig.~\ref{fig:teaserb-b}. 

To catalyze further exploration to advertising banner creation, we introduce BannerRequest400, a dedicated benchmark featuring 100 uniquely created logos paired with 400 diverse banner requests spanning multiple target audiences and primary purposes. Both quantitative and qualitative assessments confirm BannerAgency's capabilities in generating high-quality banner designs that effectively adapt to various requirements while offering unparalleled editing flexibility through its component-based output format.

Our contributions can be summarized as follows:

\begin{itemize}

    \item We propose BannerAgency, a training-free MLLM agent system that processes minimal multimodal inputs (brand logos, advertiser requirements) and autonomously generates all necessary creative assets as harmonized banner designs. 
    
    \item Simulating professional design workflows---from strategy development to technical implementation, BannerAgency leverages MLLMs' extensive world knowledge to navigate the complex design space while ensuring complete editability through component-based implementation in design tools such as Figma or as SVG format.

    \item To catalyze research in agent-driven advertising banner design, we introduce BannerRequest400, a curated benchmark featuring 100 uniquely created logos paired with 400 diverse banner requests spanning multiple target audiences and primary purposes.

    \item Through comprehensive evaluation, we exhibit BannerAgency's effectiveness in generating high-quality, versatile, and editable banner designs that can serve either as final outputs or as starting points for subjective refinement.

\end{itemize}

\section{Related Work}
\subsection{Graphic Design Generation}
Current approaches to graphic design generation face significant limitations. Pixel-based text-to-image models~\cite{dalle3,flux,stablediffusion,controlnet} require cumbersome editing, while layout generation research~\cite{zheng2019content,o2014learning,lee2020neuraleccv,lin2024spot,flexdm,layoutdm,vlc,graphist,lin2023autoposter,iwai2024layout,weng2024desigen,yun2025designlab} has not adequately addressed simultaneous element generation and arrangement~\cite{markupdm}. Existing solutions in prompt-based poster generation~\cite{cole,opencole,prompt2poster} rely on domain-specific training, accept only unimodal text input (not multimodal inputs like logos), and lack flexibility. While LayoutPrompt~\cite{layoutprompter} offers a training-free approach, it addresses only layout without end-to-end capabilities. We propose a training-free agent framework that leverages MLLMs to simultaneously generate and arrange design elements in editable code representation from multimodal inputs, enabling flexible banner creation without domain-specific fine-tuning.

\subsection{Multimodal LLM Agents}
MLLMs~\cite{mllm1,mllm2,mllm3,mllm4} have recently spurred an emerging wave of multimodal agent research~\cite{mmagent_survey1,mmagent_survey2,mmagent_survey3,lin2024battleagent,li2024mmedagent,web1}. These agents—autonomous entities that leverage tool-calling capabilities, planning, memory, and goal-directed reasoning—have been applied to various design domains including presentation slides generation~\cite{pptagent,autopresent}, image manipulation~\cite{genartist,idea2img}, infographic design~\cite{infographic1,infographic2}, and sketch creation~\cite{sketchagent}. Our work extends this evolving frontier by leveraging MLLMs' emergent capabilities to pioneer a comprehensive banner design solution that generates high-quality designs adaptable to diverse banner requests while ensuring strong editability through its component-based approach.

\section{BannerAgency}
Taking a festival promotion banner request for example, Fig.\ref{fig:overview} illustrates the overall workflow of BannerAgency. Simulating how graphic designers work in banner ad industry, our BannerAgency consists of a \textit{Strategist} to communicate with advertisers on their requirements, a \textit{Background Designer} to draw the visual canvas, a \textit{Foreground Designer} to craft the foreground elements such as logo, text, CTA button, and decorative elements, and a \textit{Developer} to render everything. Each agent is empowered by an MLLM backbone and can access existing memory and call external tools when needed.

\subsection{Strategist}
The \textit{Strategist} agent interfaces with advertisers to ground requirements for banner creation. Advertisers typically provide brand guidelines and specifications (\textit{e.g.}, logo, desired dimensions). This agent determines key banner objectives including mood, tone, target audience, and primary purpose. To maintain brand identity, the \textit{Strategist} analyzes the provided logo, trimming transparent padding to optimize space utilization while preserving visual content. Upon collecting all requirements, the Strategist stores this information in memory and transfers control to the \textit{Background Designer} agent for the next phase of banner generation.

\begin{figure}[b]
\centering     
\includegraphics[width=0.8\linewidth]{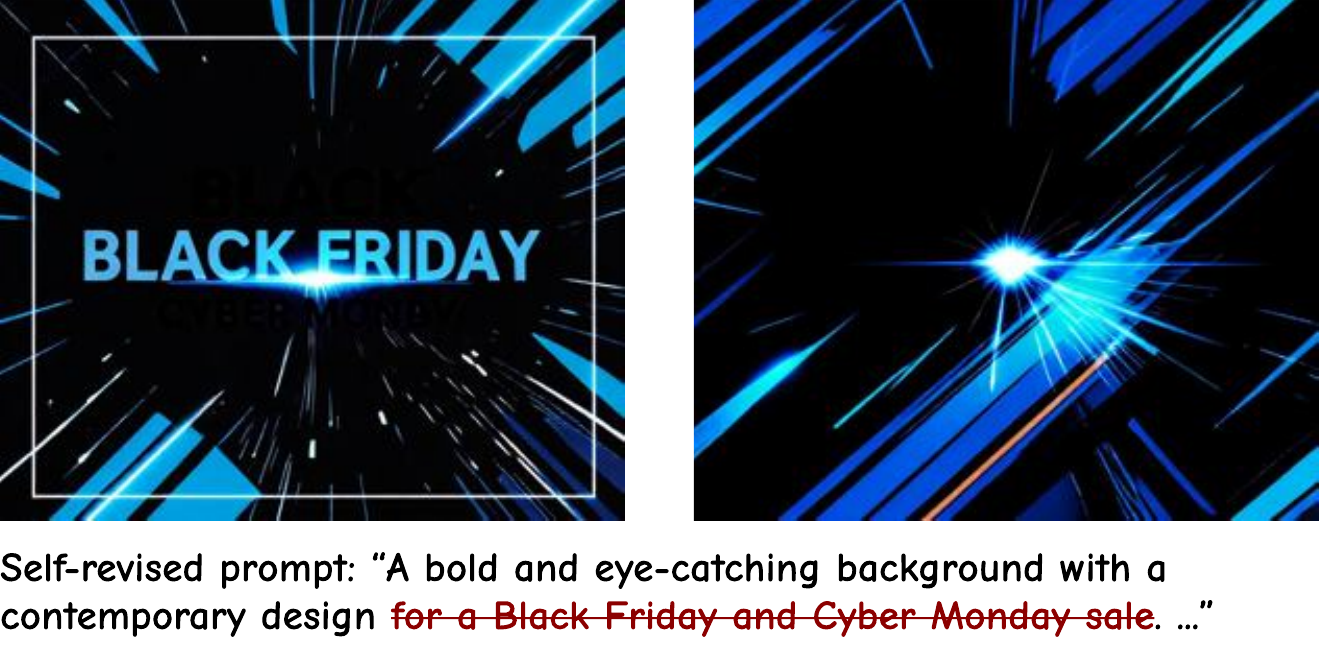}
\caption{Prompt refinement for text-free backdrop creation.}
\label{fig:bd_explain}
\end{figure}

\subsection{Background Designer}
The \textit{Background Designer} generates prompts for text-to-image (\texttt{T2I}) tools to create the visual canvas for banner ads. Implemented as a ReAct agent~\cite{react}, it harmonizes advertiser requests, logo characteristics, and campaign objectives stored in memory to produce appropriate background visuals. The agent utilizes three specialized tools: \texttt{FindImagePath} to check for existing background images provided by advertisers, \texttt{T2I} to generate new visuals, and \texttt{TextChecker} to check if the generated image contains text. For new image generation, the agent employs a self-refinement loop~\cite{selfrefine,idea2img} to ensure text-free backgrounds, as text elements in the background layer would complicate the \textit{Foreground Designer}'s work and potentially introduce difficult-to-remove gibberish text rendered directly into pixels. When text is detected by the \texttt{TextChecker} tool, the agent revises its prompt to discourage text generation and repeats the process for up to five iterations, as illustrated in Fig.\ref{fig:bd_explain} where a "Black Friday" background was successfully made text-free. During image generation, the agent selects the closest supported aspect ratio from the limited set available in the \texttt{T2I} tool that exceeds the target dimensions—ensuring higher initial resolution—and subsequently downscales to the target banner dimensions. This approach preserves visual fidelity while ensuring exact sizing for various display contexts. The resulting text-free background serves as an appropriate visual foundation, reserving space for the \textit{Foreground Designer} to add text, CTA buttons, logos, and decorative elements in subsequent stages.

\begin{figure}[t]
\centering     
\includegraphics[width=\linewidth]{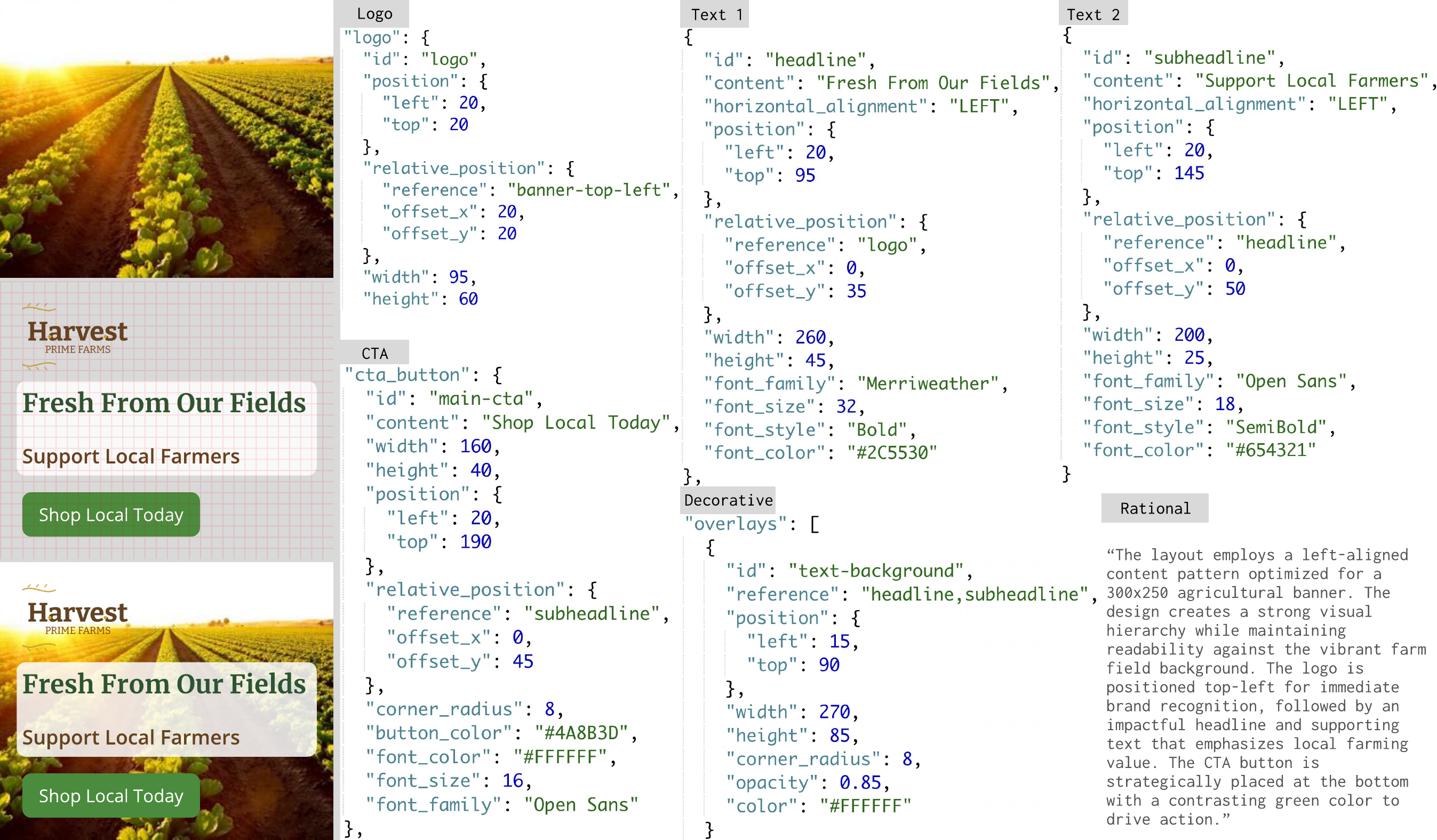}
\caption{Element-wise blueprint as \textit{Foreground Designer} output.}
\label{fig:fd_format}
\end{figure}

\subsection{Foreground Designer}
The \textit{Foreground Designer} agent serves as the creative cornerstone of our system, generating a structured blueprint for all foreground elements encoded as a JSON-structured schema---defining precise positioning, styling, and content for logos, campaign texts, call-to-action buttons, and decorative elements. To maintain good alignment and overlap, we use relative positioning with reference properties to model the spatial relationship among banner components. An example of this element-wise blueprint from \textit{Foreground Designer} is shown in Fig.~\ref{fig:fd_format}. By transforming design decisions into structured data rather than rasterized pixels, the \textit{Foreground Designer} creates a foundation for both visual coherence and complete element-level editability.

\begin{figure}[t]
\centering     
\includegraphics[width=\linewidth]{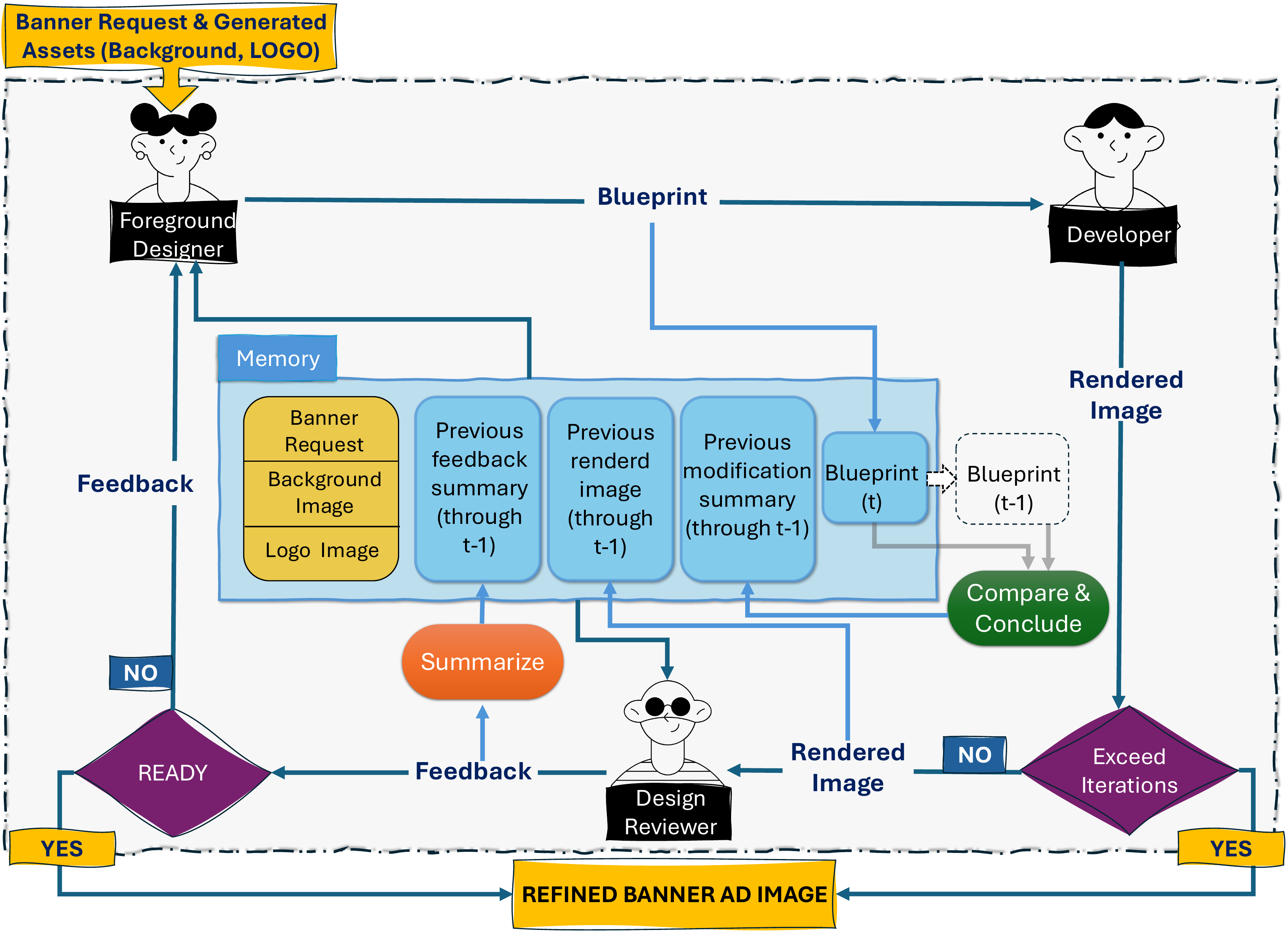}
\caption{Memory-augmented iterative design refinement process.}
\label{fig:refine_illust}
\end{figure}

\paragraph{Memory-augmented iterative design refinement.}
The evolution of designs through iterative external feedback and refinement is a staple procedure in professional design~\cite{adcreative2024}. Simulating this practice~\cite{selfrefine,idea2img}, our memory-augmented iterative design refinement process, illustrated in Fig.~\ref{fig:refine_illust}, enhances banner ad designs by introducing an external critic \textit{Design Reviewer}. At initial creation ($t=0$), the \textit{Foreground Designer} generates the first blueprint based solely on the creative brief (banner request, background image, logo). The \textit{Developer} renders this blueprint, and the \textit{Design Reviewer} provides initial feedback, which is summarized and stored in memory. For the first refinement ($t=1$), the \textit{Foreground Designer} refines the blueprint based on this feedback, comparing the new blueprint with the previous one and summarizing into a modification list (\textit{ie}, "Compare \& Conclude''). For subsequent refinements ($t>1$), both designer and reviewer access the memory that is continuously augmented with new feedback, renderings, and modification summaries after each iteration, enabling increasingly informed decisions. This memory-augmented approach, where each iteration enriches the shared memory with new design artifacts and critiques, ensures both designer and reviewer maintain awareness of design evolution while continuously aligning with brand objectives. The process concludes when iterations are exceeded or the design is deemed production-ready by the reviewer, resulting in the final refined banner ad image.

\subsection{Developer}
The \textit{Developer} plays a crucial role in converting the \textit{Foreground Designer}'s blueprint into a banner ad image. We propose two implementations for the \textit{Developer}: generating SVG code or Figma plugin code. Examples of codes in two formats are included in Appendix Fig.~\ref{fig:code_format}.

\paragraph{SVG Code Generation.}
Scalable Vector Graphics (SVG) is a widely-supported format for rendering two-dimensional graphics in web browsers and design software~\footnote{https://www.w3.org/TR/SVG11/}. It is a declarative XML-based markup language with elements that directly define visual properties through coordinates, dimensions, and styling attributes. This approach provides flexibility and compatibility, as SVG is an open standard supported by various design tools. However, it requires an additional step of importing the SVG code into a design editing software such as Adobe Illustrator or Inkscape, and offers a more static representation of the design.

\paragraph{Figma Plugin Code Generation.}
Figma is a popular cloud-based design and prototyping tool used by designers worldwide~\footnote{https://www.figma.com/}. By translating the blueprint into JavaScript code that adheres to the Figma Plugin API~\footnote{https://www.figma.com/plugin-docs/}, the \textit{Developer} enables seamless integration with the Figma ecosystem. The generated code includes imperative programming instructions to programmatically create and manipulate nodes in the Figma document hierarchy. This approach builds the design through a sequence of operations rather than a static description. The advantage of this approach is that the banner ad image can be automatically rendered within Figma, eliminating the need for manual importing. This streamlines the design workflow and allows for quick previews and iterations while leveraging Figma's native component system and interactive editing capabilities.

\begin{figure}[t]
  \centering
  \begin{subfigure}{0.45\linewidth}
    {\includegraphics[width=\linewidth]{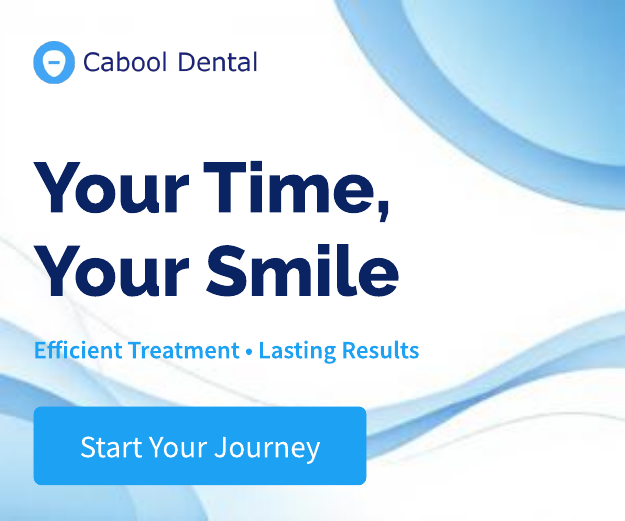}}
    \caption{Working professionals.}
    \label{fig:workingp}
  \end{subfigure}
  \hfill
  \begin{subfigure}{0.45\linewidth}
    {\includegraphics[width=\linewidth]{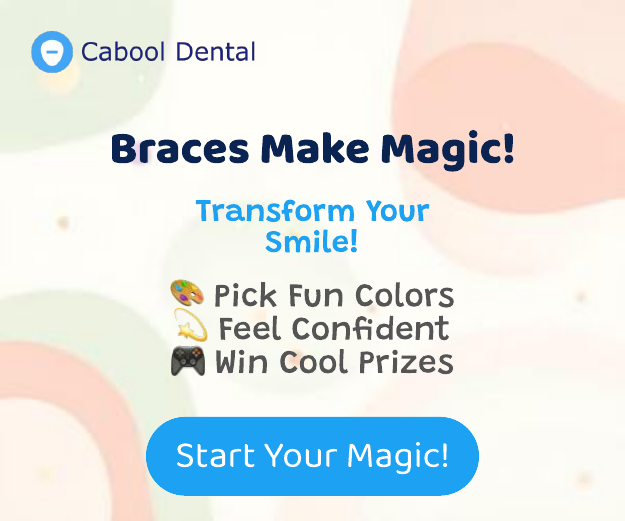}}
    \caption{Kids.}
    \label{fig:kids}
  \end{subfigure}
  \caption{Same campaign for different audiences.}
  \label{fig:ds_audience}
\end{figure}

\section{BannerRequest400 Benchmark}

Existing design datasets, such as DESIGNERINTENTION~\cite{opencole}, are limited to uni-modal designer intentions in text format and lack specific addressing of banner advertisement requirements. To address this gap, we introduce BannerRequest400, the first multimodal benchmark specifically designed for evaluating advertising banner generation systems. This novel benchmark pairs brand logos (visual modality) with corresponding banner design requests (textual modality). We curated 100 advertising design intentions from DESIGNERINTENTION and synthesized corresponding vector logos using Claude3.5 Sonnet~\cite{claude}, followed by expert review and refinement to visual fidelity, avoid bias, and maintain authentic logo aesthetics. Our empirical analysis, illustrated in Fig.\ref{fig:ds_audience}, demonstrates how identical dental clinic promotion requests produce substantially different designs based on target audience---with professional-oriented banners emphasizing efficiency and results, while child-focused versions feature playful elements and incentives. Leveraging this insight, we employed GPT-4o\cite{gpt4o} to systematically expand each intention into four distinct banner requests, each targeting a unique audience-purpose combination, resulting in 400 diverse specifications. To ensure industry relevance, we further extended these 400 requests across 13 standard banner dimensions, yielding a comprehensive evaluation set of 5,200 multimodal banner specifications. For benchmarking text-to-image generation methods, we further enhanced each request with detailed specifications that incorporate logo-specific visual characteristics using GPT-4o. BannerRequest400 enables the first rigorous evaluation of banner generation approaches across diverse design requests, facilitating advancement in automatic advertising banner design generation.

\section{Experiments}

\subsection{Experimental Setup}

\paragraph{Algorithms.} For baseline approaches on our BannerRequest400 benchmark, we compare with two powerful pixel-based image generation models including closed-sourced DALL-E3~\cite{dalle3} and open-sourced FLUX.1-schnell~\cite{flux} which is also our backbone for the \texttt{T2I} tool. We also compare with OpenCOLE~\cite{cole,opencole}, a prompt-based graphic design model that relies on fine-tuning. For the backbone of our proposed training-free BannerAgency, we explore two multimodal LLM variants including GPT-4o (2024-10-21)~\cite{gpt4o} and Claude3.5-Sonnet (2024-10-22)~\cite{claude} with temperature set as 0.3. We provide more details including all agent profile definitions and tool definitions in Appendix~\ref{sec:setup}.
\paragraph{Metrics.} 
To evaluate banner design quality, we developed six metrics (Tab.~\ref{tbl:metricsdef}) from extensive literature survey on banner design principles~\cite{metricsurvey1,metricsurvey2,metricsurvey3,metricsurvey4,metricsurvey5,metricsurvey6}. These metrics capture advertising-specific aspects overlooked by general design metrics~\cite{opencole,gptevaldesign}: audience alignment, brand integration, call-to-action effectiveness, message clarity, brand consistency, and visual appeal---addressing both marketing functionality and aesthetic requirements critical for digital advertising. For completeness and comparison with prior metrics, we also include results using general design metrics in Appendix~\ref{sec:ra-b}.

\begin{table}[t]
\centering
\LARGE
\resizebox{\columnwidth}{!}{
\begin{tabular}{p{0.1\textwidth} p{0.8\textwidth}lll} 
\toprule 
{\bf Metric} & \multicolumn{1}{c}{\bf Criteria} & \bf Pearson & \bf Spearman & \bf ICC  \\
\midrule
{TAA} & {Measures how well the generated banner ad aligns with the given request, including the theme, target audience, and primary purpose.} & 0.854 & 0.840 & 0.922 \\
\midrule
{LPS} & {Evaluates whether the logo is well-integrated into the design in terms of visibility, size, and positioning.} & 0.955 & 0.960 & 0.986 \\
\midrule
{CTAE} & {Evaluates whether the Call-to-Action (CTA) is clear, engaging, and visually emphasized. } & 0.927 & 0.946 & 0.947 \\
\midrule
{CPYQ} & {Evaluates the effectiveness of the headline, subheadline, and any other text in the banner ad, focusing on clarity, readability, persuasiveness, and grammatical correctness.} & 0.856 & 0.840 & 0.962\\
\midrule
{BIS}  & {Measures how well the banner ad visually and stylistically aligns with the brand’s identity beyond just logo placement. This includes color consistency, typography, imagery, and overall brand feel.} & 0.877 & 0.871 & 0.935 \\
\midrule
{AQS} & {Measures the visual appeal, including color harmony, layout balance, typography, and overall design quality.} & 0.892 & 0.895 & 0.973\\
\bottomrule 
\end{tabular}
}
\vspace{-2mm}
\caption{\textbf{Banner design evaluation metrics and human-LLM correlation.} Each metric (rated 1-5) assesses a distinct aspect of banner design quality. Pearson, Spearman, and ICC correlation coefficients between human and GPT-4o ratings show strong agreement (all p $<$ 0.001). TAA: Target Audience Alignment; LPS: Logo Placement Score; CTAE: CTA Effectiveness; CPYQ: Copywriting Quality; BIS: Brand Identity Score; AQS: Aesthetic Quality Score. Detailed rating criteria are provided in supplementary.
}
\vspace{-2mm}
\label{tbl:metricsdef}
\end{table}
\paragraph{Human study.}
We conducted three human evaluations to assess different aspects of BannerAgency. First, a system preference study with 20 participants comparing banner designs generated by five BannerAgency variants across 20 randomly selected requests through pairwise comparisons. Second, a refinement effectiveness evaluation where 15 participants assessed quality progression across four iterations of 20 different banner designs, rating both initial and best versions on a 5-point scale. Third, a human-LLM correlation study validating the alignment between GPT-4o's automated scoring and human perception across six proposed metrics, with 17-19 participants independently grading 25 representative images per metric. See Appendix~\ref{sec:hs} for details.

\begin{table*}[t]
\footnotesize
\centering
    \resizebox{0.5\linewidth}{!}{
    \begin{tabular}{lllllll}
    \toprule
    \textbf{Method}           & \textbf{TAA} & \textbf{LPS} & \textbf{CTAE} & \textbf{CPYQ} & \textbf{BIS}  & \textbf{AQS}  \\
    \hline
    \rowcolor{yellow!15}
    \multicolumn{7}{c}{\textit{Baselines}} \\
    \hline
    DALL-E3*       &   3.17      & 1.78   & 4.18  & 2.48  & 3.06    &  3.10 \\
    FLUX.1-schnell*    &   3.36       &  3.67  & 4.28  & 2.64  & 3.09   &  3.12             \\
    OpenCOLE    &    3.56       & 1.20  & 2.85  & 3.20  & 2.84      &   3.29              \\
    \hline
    \rowcolor{yellow!15}
    \multicolumn{7}{c}{\textit{BannerAgency}} \\
    \hline
\multicolumn{7}{l}{\quad\textit{SVG Implementation}} \\
\makebox[0.5cm][l]{$\mathbf{[E]}$} GPT-4o w/o BG & 4.15 & 4.51 & 4.04 & 4.25 & 3.48 & 3.19 \\
\makebox[0.5cm][l]{$\mathbf{[D]}$} Claude3.5-Sonnet w/o BG   & 4.40 & \textbf{4.57} & 4.63 & 4.48 & 4.26 & 3.77 \\

\makebox[0.5cm][l]{$\mathbf{[A]}$} Claude3.5-Sonnet & \textbf{4.56} & 4.43 & \textbf{4.94} & \textbf{4.53} & \textbf{4.36} & \textbf{3.92} \\
\cmidrule(l){1-7}
\multicolumn{7}{l}{\quad\textit{Figma Implementation}} \\
\makebox[0.5cm][l]{$\mathbf{[C]}$} GPT-4o & \textbf{4.39} & 4.22 & 4.66 & 4.00 & 4.14 & 3.69 \\
\makebox[0.5cm][l]{$\mathbf{[B]}$} Claude3.5-Sonnet & 4.34 & \textbf{4.33} & \textbf{4.81} & \textbf{4.19} & \textbf{4.22} & \textbf{3.85} \\
\bottomrule
  \end{tabular}
  \label{tab:quant_metrics}
    }
    \blank{0.05cm}
    \resizebox{0.45\linewidth}{!}{
    \begin{tabular}{c}
    \includegraphics[width=1\linewidth]{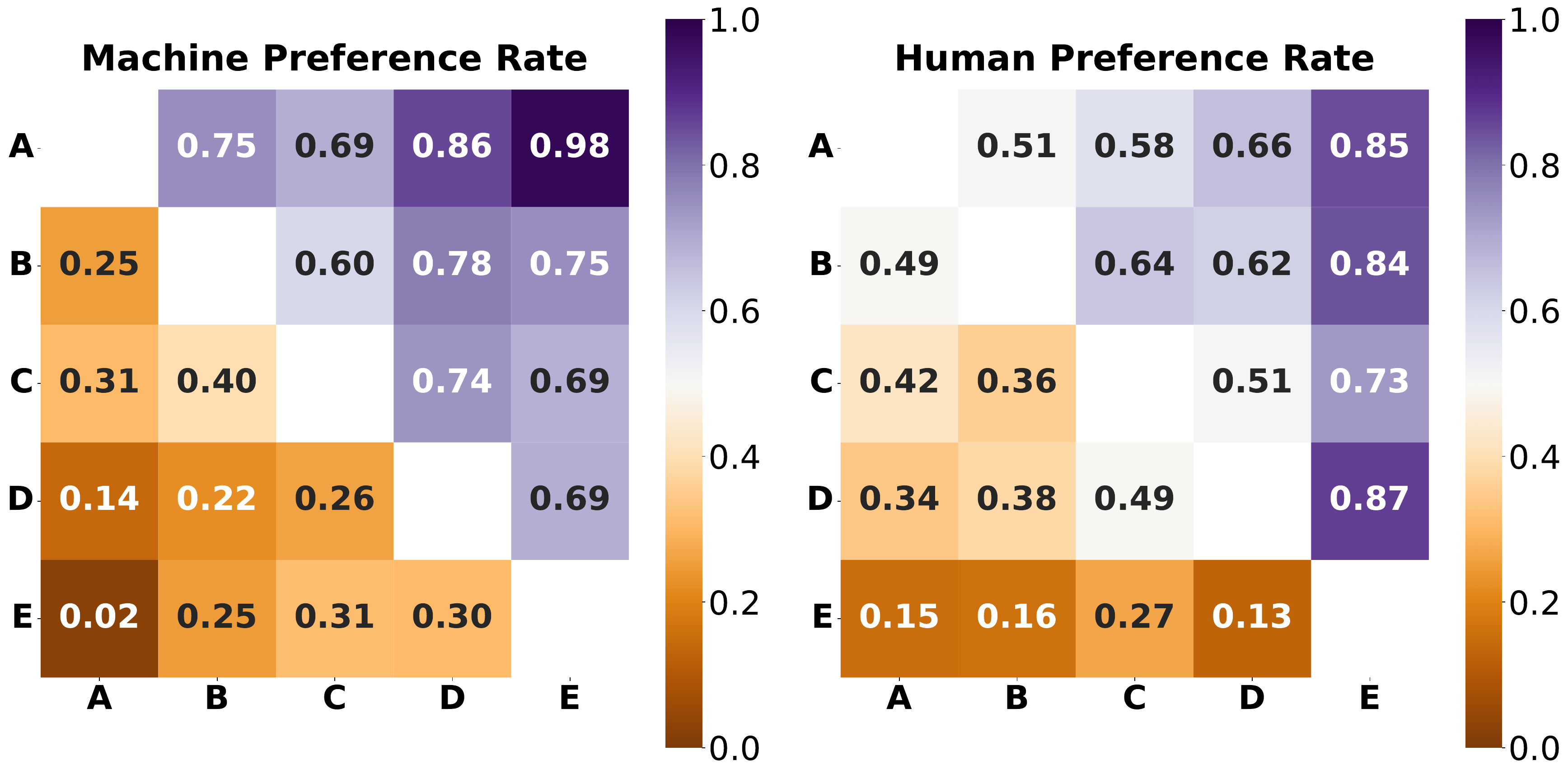}
    \label{tab:quant_preference}
    \end{tabular}}
    \\
    (a) Banner quality metrics (GPT-4o, human-validated in Tab.~\ref{tbl:metricsdef}). \blank{1.3cm} (b) System preference rates by GPT-4o and humans.\blank{0.1cm}
    \vspace{-2mm}
   
        \caption{\textbf{Quantitative comparison on BannerRequest400 benchmark.}}
     \vspace{-3mm}
\label{tab:quant}
\end{table*}

\begin{figure*}[t]
\centering     
\includegraphics[width=\linewidth]{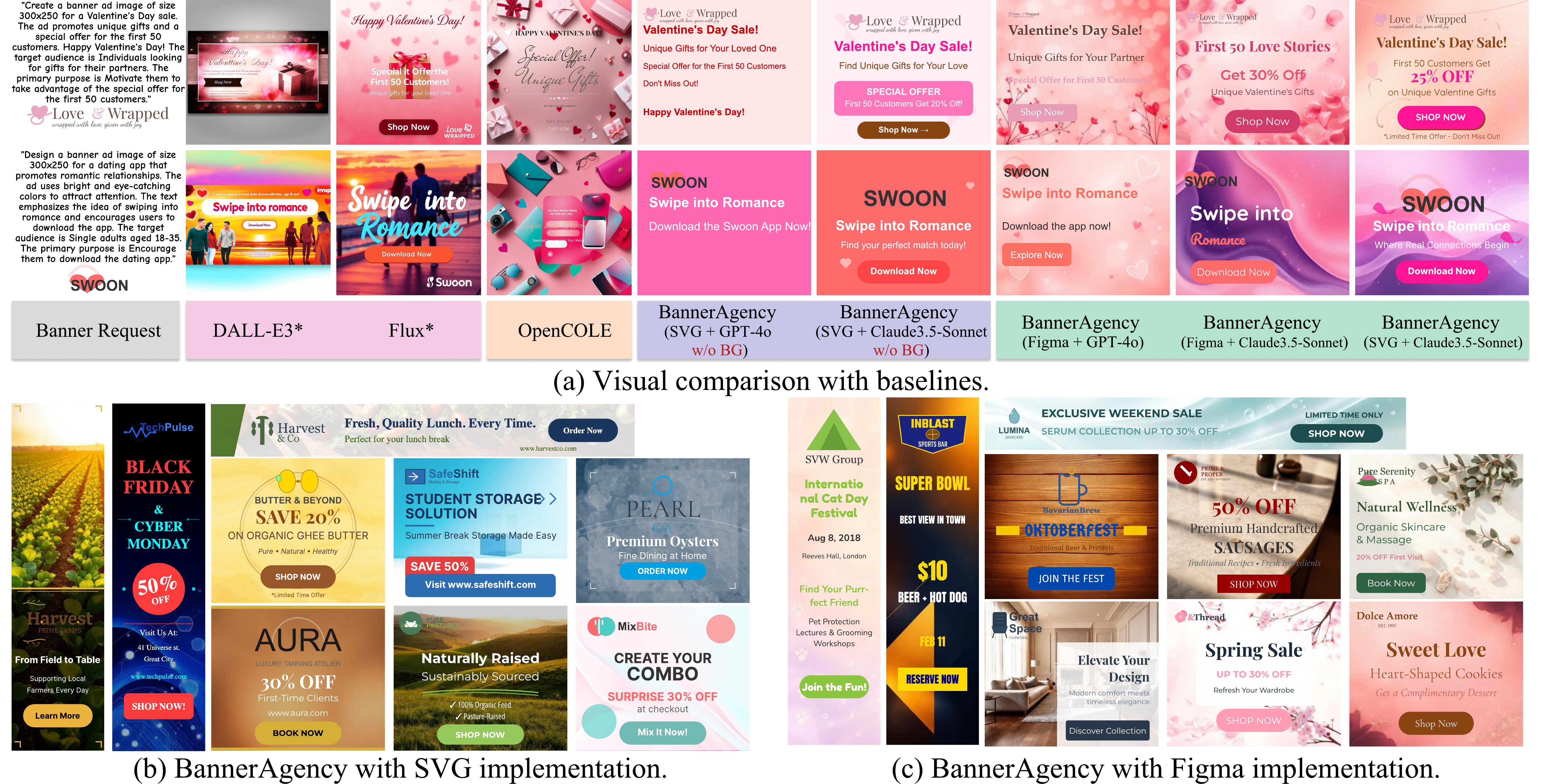}
\caption{\textbf{Visual results.} * denotes methods requiring explicit banner descriptions, whose detailed prompts are provided in Appendix~\ref{sec:dp}. More visual results can be found in Appendix~\ref{sec:moreresults}.}
\label{fig:visual_compare}
\end{figure*}



\begin{figure*}[t]
\centering     
\includegraphics[width=\linewidth]{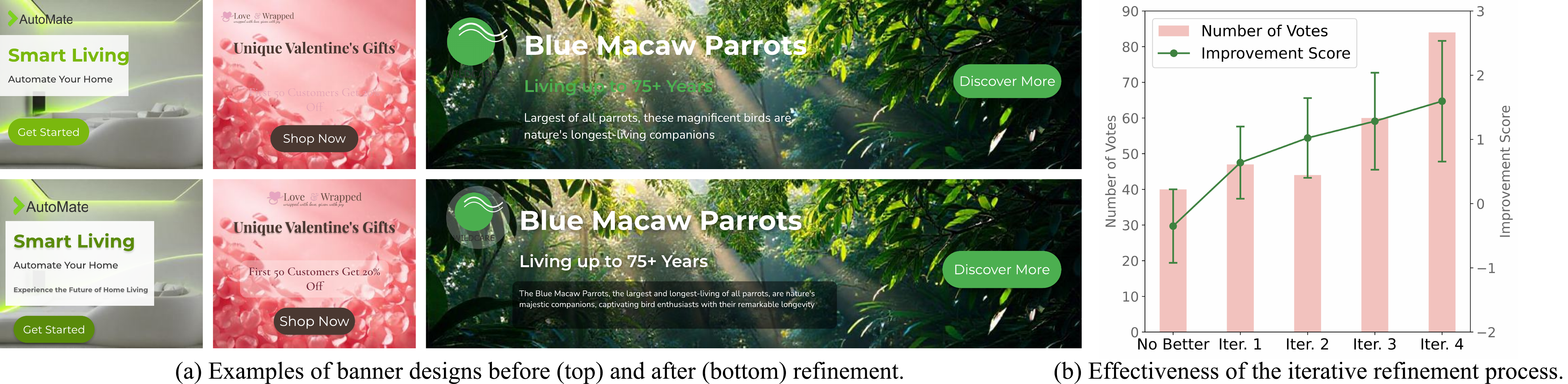}
\caption{\textbf{BannerAgency's design refinement.} Human study reveals later iterations received more votes (pink bars) with fourth iteration most preferred. Green line shows mean improvement between initial and selected versions, demonstrating significant quality gains.}
\label{fig:refine_exp}
\end{figure*}


\begin{figure*}[t]
\centering     
\includegraphics[width=\linewidth]{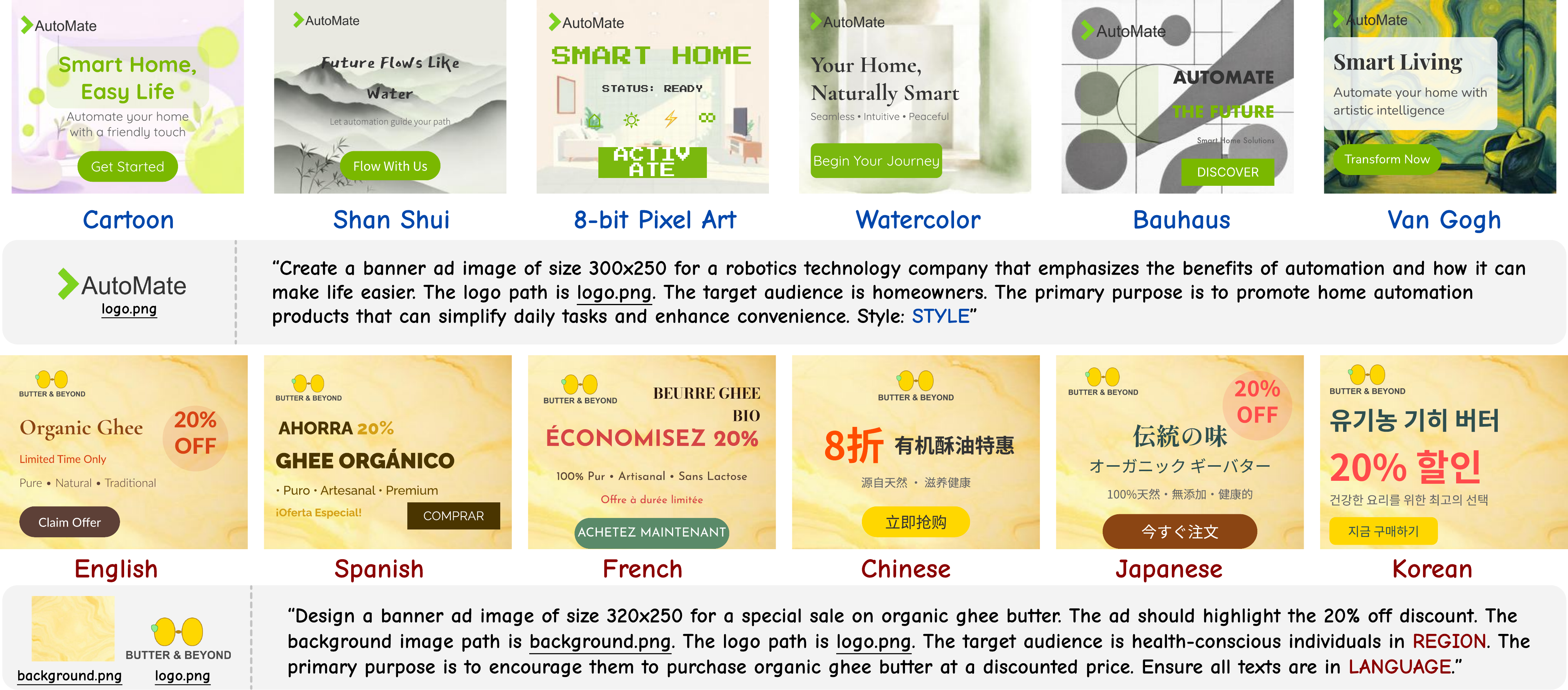}
\caption{\textbf{BannerAgency's design versatility.} We demonstrate the high versatility through cross-genre aesthetic adaptation (top), and cultural adaptation (bottom) across diverse market contexts, all without requiring additional training or fine-tuning.}
\label{fig:control}
\end{figure*}


\subsection{Results and Analysis}
\paragraph{Comparison with baselines.}
We present quantitative and visual comparisons in Tab.~\ref{tab:quant} (a) and Fig.~\ref{fig:visual_compare}, respectively. BannerAgency outperforms all baselines across all aspects. Baseline limitations include: fixed-size generation preventing non-standard dimensions like $300\times250$, failure to incorporate provided logos, and incorrect text rendering requiring correction~\cite{typer}. Among BannerAgency variants, our preference heatmap in Tab.~\ref{tab:quant} (b) shows Claude3.5-Sonnet surpasses GPT-4o ($\mathbf{[D]}$ vs. $\mathbf{[E]}$ and $\mathbf{[B]}$ vs. $\mathbf{[C]}$), especially in SVG implementation. Human evaluators showed no preference between SVG and Figma implementations ($\mathbf{[A]}$ vs. $\mathbf{[B]}$). Notably, removing the Background Designer component (\textit{ie}, w/o BG) significantly impairs visual appeal, resulting in plain and unengaging designs, particularly with the GPT-4o backbone. This underscores the critical role of specialized background generation in creating visually compelling banner advertisements.

\paragraph{Refinement process.}
Fig.~\ref{fig:refine_exp} (a) demonstrates the design improvement after the iterative refinement process. The initial designs are upgraded by enhancing content clarity, layout composition, and visual styling without radical redesigns. As shown in Fig.~\ref{fig:refine_exp} (b), our evaluation reveals a clear upward trend in quality, with improvement scores steadily increasing across iterations and later versions receiving more votes. Notably, the vote distribution indicates that refinement effectiveness is design-context dependent. While the final iteration was most frequently preferred, many participants favored earlier iterations or the initial design. This finding highlights the value of maintaining a complete version track, allowing advertisers to select the specific iteration that best aligns with their vision, regardless of where it falls in the refinement sequence. This process also reflects how our component-based approach enables practical, editable design alternatives. The full refinement process is visualized in Appendix~\ref{sec:ra-a}.

\paragraph{Versatility to user prompts.}
We showcase in Fig.~\ref{fig:control} two more examples highlighting cross-genre aesthetic adaptation and cross-cultural linguistic flexibility, respectively. We observe how the designs adapt coherently to each style while maintaining color schemes that complement the AutoMate logo. For different linguistic requests with same canvas input, we show how the system successfully accommodates language-specific conventions and cultural preferences with tailored fonts and styles, which directly addresses the industry challenge of creating culturally appropriate marketing materials without specialized design teams. Empowered by the world knowledge of multimodal LLMs, our training-free BannerAgency demonstrates remarkable adaptability to diverse user requirements without any additional fine-tuning or domain-specific datasets.

\section{Conclusion}
We introduced BannerAgency, a training-free MLLM agent system that transforms advertising banner design by simulating professional design teams through specialized agents (\textit{Strategist}, \textit{Background Designer}, \textit{Foreground Designer}, and \textit{Developer}). Our approach generates component-based, editable designs in industry-standard formats from multimodal inputs without domain-specific fine-tuning, overcoming the limitations of pixel-based methods. Through our BannerRequest400 benchmark featuring 100 unique logos paired with 400 diverse banner requests, we demonstrate BannerAgency's effectiveness in producing high-quality, versatile, and editable designs that kickstart the creative design process. This represents a significant step toward democratizing professional-grade design capabilities across the digital advertising landscape. 

\section*{Limitations}
While our approach represents a significant step toward leveraging MLLM agents for versatile and editable graphic design generation, there exist some limitations. First, we exclusively experiment with logo images as multimodal input. It is sufficient for abstract service banner ads but neglects product images which are also crucial in advertisements. Second, the decorative elements are limited to simple geometric shapes due to the lack of finetuning or domain-specific guidance. Similarly, we do not address visual texts with artistic effect. These limitations reflect our primary objective: to validate design knowledge already embedded in MLLMs and demonstrate the agent system's versatility across varying requests (banner sizes, styles, etc.) while maintaining output editability. In the future, we will dedicate efforts to the upgrade of each agent by equipping them with more advanced tools or further decompose different agents into hierarchical teams. To reiterate, we aim to extend BannerAgency by: (1) expanding multimodal input to incorporate product images, (2) introducing more sophisticated decorative elements beyond basic geometric shapes, (3) allowing for more hierarchical architecture for better collaboration among agents, and (4) considering conversion metric optimization to directly connect design choices with advertising performance.

\section*{Ethics Statement}
The BannerAgency framework, while advancing automated banner design through MLLM agents that simulate professional design workflows, acknowledges important ethical considerations. We recognize potential concerns regarding employment impacts on human designers, conscious content generation that avoids harmful stereotypes or misleading advertising, intellectual property considerations when incorporating brand assets, transparency with end users about AI-generated content, and potential biases in design choices that may perpetuate inequities in representation. Our component-based approach deliberately preserves human creative oversight through editable outputs, ensuring designers maintain agency in the final creative process rather than being replaced. We commit to ongoing evaluation of these ethical dimensions, including regular bias assessments of generated content, and encourage the research community to further explore responsible deployment strategies for automated design systems that complement rather than supplant human creativity.

\bibliography{custom}

\begin{thebibliography}{87}
\providecommand{\natexlab}[1]{#1}

\bibitem[{Abdin et~al.(2024)Abdin, Aneja, Awadalla, Awadallah, Awan, Bach, Bahree, Bakhtiari, Bao, Behl et~al.}]{mllm3}
Marah Abdin, Jyoti Aneja, Hany Awadalla, Ahmed Awadallah, Ammar~Ahmad Awan, Nguyen Bach, Amit Bahree, Arash Bakhtiari, Jianmin Bao, Harkirat Behl, and 1 others. 2024.
\newblock Phi-3 technical report: A highly capable language model locally on your phone.
\newblock \emph{arXiv preprint arXiv:2404.14219}.

\bibitem[{ADCreative.ai(2024)}]{adcreative2024}
ADCreative.ai. 2024.
\newblock \href {https://www.adcreative.ai/post/the-best-kept-secrets-about-the-iterative-design-process} {The best kept secrets about the iterative design process}.
\newblock Accessed: 2025-02-28.

\bibitem[{Alibaba(2020)}]{luban}
Alibaba. 2020.
\newblock \href {https://www.alibabacloud.com/blog/the-evolution-of-luban-in-designing-one-billion-images_596118} {The evolution of luban in designing one billion images}.
\newblock Accessed: 2025-02-28.

\bibitem[{Anthropic(2024)}]{claude}
Anthropic. 2024.
\newblock \href {https://www.anthropic.com/news/claude-3-5-sonnet} {Claude 3.5 sonnet}.

\bibitem[{Bannerwise(2022)}]{bannersize}
Bannerwise. 2022.
\newblock \href {https://www.bannerwise.io/blog/understanding-ad-sizes/?gad_source=1&gclid=Cj0KCQiAz6q-BhCfARIsAOezPxmPtlEB-bNGRFYk-yPLcDY9wQp3K2q2KNt66xZNU0dzdxOrpgPikoYaAqduEALw_wcB} {Does size matter? understanding ad sizes and when to use them}.
\newblock Accessed: 2025-02-28.

\bibitem[{Betker et~al.(2023)Betker, Goh, Jing, Brooks, Wang, Li, Ouyang, Zhuang, Lee, Guo et~al.}]{dalle3}
James Betker, Gabriel Goh, Li~Jing, Tim Brooks, Jianfeng Wang, Linjie Li, Long Ouyang, Juntang Zhuang, Joyce Lee, Yufei Guo, and 1 others. 2023.
\newblock Improving image generation with better captions.
\newblock \emph{OpenAI}.

\bibitem[{BlackForestLab(2024)}]{flux}
BlackForestLab. 2024.
\newblock \href {https://blackforestlabs.ai/announcing-black-forest-labs/} {Announcing black forest labs}.

\bibitem[{Canva(2025)}]{canvabanner}
Canva. 2025.
\newblock \href {https://www.canva.com/create/banners/} {Free banner maker}.
\newblock Accessed: 2025-02-28.

\bibitem[{Chen and Huang(2024)}]{infographic2}
Chao-Ting Chen and Hen-Hsen Huang. 2024.
\newblock Integrating llm, vlm, and text-to-image models for enhanced information graphics: A methodology for accurate and visually engaging visualizations.
\newblock In \emph{IJCAI}, pages 8627--8630.

\bibitem[{Chen et~al.(2021)Chen, Xu, Jiang, Ge, Zhang, Lian, and Zheng}]{chen2021automated}
Jin Chen, Ju~Xu, Gangwei Jiang, Tiezheng Ge, Zhiqiang Zhang, Defu Lian, and Kai Zheng. 2021.
\newblock Automated creative optimization for e-commerce advertising.
\newblock In \emph{WWW}, pages 2304--2313.

\bibitem[{Chen et~al.(2024)Chen, Huang, Lv, Cui, Chen, and Wei}]{textdiffuser}
Jingye Chen, Yupan Huang, Tengchao Lv, Lei Cui, Qifeng Chen, and Furu Wei. 2024.
\newblock Textdiffuser-2: Unleashing the power of language models for text rendering.
\newblock In \emph{ECCV}, pages 386--402. Springer.

\bibitem[{Chen et~al.(2025)Chen, Feng, Du, Wang, Chen, Wang, Liu, Li, Zhao, Li et~al.}]{chen2025ctr}
Xingye Chen, Wei Feng, Zhenbang Du, Weizhen Wang, Yanyin Chen, Haohan Wang, Linkai Liu, Yaoyu Li, Jinyuan Zhao, Yu~Li, and 1 others. 2025.
\newblock Ctr-driven advertising image generation with multimodal large language models.
\newblock In \emph{WWW}.

\bibitem[{Cheng et~al.(2024)Cheng, Zhang, Yang, Nie, Li, Wu, and Shao}]{graphist}
Yutao Cheng, Zhao Zhang, Maoke Yang, Hui Nie, Chunyuan Li, Xinglong Wu, and Jie Shao. 2024.
\newblock Graphic design with large multimodal model.
\newblock \emph{arXiv preprint arXiv:2404.14368}.

\bibitem[{Choi et~al.(2018)Choi, Aizawa, and Sebe}]{choi2018fontmatcher}
Saemi Choi, Kiyoharu Aizawa, and Nicu Sebe. 2018.
\newblock Fontmatcher: font image paring for harmonious digital graphic design.
\newblock In \emph{IUI}, pages 37--41.

\bibitem[{community(2025)}]{whystrategist}
The~LinkedIn community. 2025.
\newblock \href {https://www.linkedin.com/advice/1/how-do-you-know-your-graphics-working-marketing-campaigns} {How do you know if your graphics are working in your marketing campaigns?}
\newblock Accessed: 2025-02-28.

\bibitem[{Du et~al.(2024)Du, Feng, Wang, Li, Wang, Li, Zhang, Lv, Zhu, Jin et~al.}]{du2024towards}
Zhenbang Du, Wei Feng, Haohan Wang, Yaoyu Li, Jingsen Wang, Jian Li, Zheng Zhang, Jingjing Lv, Xin Zhu, Junsheng Jin, and 1 others. 2024.
\newblock Towards reliable advertising image generation using human feedback.
\newblock In \emph{ECCV}, pages 399--415. Springer.

\bibitem[{Durante et~al.(2024)Durante, Huang, Wake, Gong, Park, Sarkar, Taori, Noda, Terzopoulos, Choi et~al.}]{mmagent_survey1}
Zane Durante, Qiuyuan Huang, Naoki Wake, Ran Gong, Jae~Sung Park, Bidipta Sarkar, Rohan Taori, Yusuke Noda, Demetri Terzopoulos, Yejin Choi, and 1 others. 2024.
\newblock Agent ai: Surveying the horizons of multimodal interaction.
\newblock \emph{arXiv preprint arXiv:2401.03568}.

\bibitem[{Express(2025)}]{adobeexpress}
Adobe Express. 2025.
\newblock \href {https://www.adobe.com/express/create/banner/graphic} {Design graphic banners for free}.
\newblock Accessed: 2025-02-28.

\bibitem[{Fong and See(2024)}]{fong2024branddiffusion}
Bi~Qi Fong and John See. 2024.
\newblock Branddiffusion: Multimodal personalized marketing visual content generation.
\newblock In \emph{ACM MM Workshop on Multimedia Content Generation and Evaluation: New Methods and Practice}, pages 72--77.

\bibitem[{Ge et~al.(2025)Ge, Wang, Zhou, Peng, Subramanian, Tan, Sap, Suhr, Fried, Neubig et~al.}]{autopresent}
Jiaxin Ge, Zora~Zhiruo Wang, Xuhui Zhou, Yi-Hao Peng, Sanjay Subramanian, Qinyue Tan, Maarten Sap, Alane Suhr, Daniel Fried, Graham Neubig, and 1 others. 2025.
\newblock Autopresent: Designing structured visuals from scratch.
\newblock \emph{arXiv preprint arXiv:2501.00912}.

\bibitem[{Guo et~al.(2021)Guo, Jin, Sun, Li, Li, Shi, and Cao}]{vinci}
Shunan Guo, Zhuochen Jin, Fuling Sun, Jingwen Li, Zhaorui Li, Yang Shi, and Nan Cao. 2021.
\newblock Vinci: an intelligent graphic design system for generating advertising posters.
\newblock In \emph{CHI}, pages 1--17.

\bibitem[{Gupta and Kembhavi(2023)}]{gupta2023visual}
Tanmay Gupta and Aniruddha Kembhavi. 2023.
\newblock Visual programming: Compositional visual reasoning without training.
\newblock In \emph{CVPR}, pages 14953--14962.

\bibitem[{Haraguchi et~al.(2024)Haraguchi, Inoue, Shimoda, Mitani, Uchida, and Yamaguchi}]{gptevaldesign}
Daichi Haraguchi, Naoto Inoue, Wataru Shimoda, Hayato Mitani, Seiichi Uchida, and Kota Yamaguchi. 2024.
\newblock \href {https://doi.org/10.1145/3681758.3698010} {Can gpts evaluate graphic design based on design principles?}
\newblock In \emph{SIGGRAPH Asia 2024 Technical Communications}, SA '24, New York, NY, USA. Association for Computing Machinery.

\bibitem[{He et~al.(2024)He, Liu, Chen, Tian, Liu, Chi, Liu, Yuan, Xing, Wang et~al.}]{mmagent_survey3}
Yingqing He, Zhaoyang Liu, Jingye Chen, Zeyue Tian, Hongyu Liu, Xiaowei Chi, Runtao Liu, Ruibin Yuan, Yazhou Xing, Wenhai Wang, and 1 others. 2024.
\newblock Llms meet multimodal generation and editing: A survey.
\newblock \emph{arXiv preprint arXiv:2405.19334}.

\bibitem[{Hu et~al.(2021)Hu, Zhang, and Liang}]{hu2021study}
Hao Hu, Chao Zhang, and Yanxue Liang. 2021.
\newblock A study on the automatic generation of banner layouts.
\newblock \emph{Computers \& Electrical Engineering}, 93:107269.

\bibitem[{Hu et~al.(2024)Hu, Iscen, Jain, Kipf, Yue, Ross, Schmid, and Fathi}]{hu2024scenecraft}
Ziniu Hu, Ahmet Iscen, Aashi Jain, Thomas Kipf, Yisong Yue, David~A Ross, Cordelia Schmid, and Alireza Fathi. 2024.
\newblock Scenecraft: An llm agent for synthesizing 3d scenes as blender code.
\newblock In \emph{ICML}.

\bibitem[{Huang et~al.(2024)Huang, Lu, Lanir, Lischinski, Cohen-Or, and Huang}]{infographic1}
Qirui Huang, Min Lu, Joel Lanir, Dani Lischinski, Daniel Cohen-Or, and Hui Huang. 2024.
\newblock Graphimind: Llm-centric interface for information graphics design.
\newblock \emph{arXiv preprint arXiv:2401.13245}.

\bibitem[{Iluz et~al.(2023)Iluz, Vinker, Hertz, Berio, Cohen-Or, and Shamir}]{iluz2023word}
Shir Iluz, Yael Vinker, Amir Hertz, Daniel Berio, Daniel Cohen-Or, and Ariel Shamir. 2023.
\newblock Word-as-image for semantic typography.
\newblock \emph{TOG}, 42(4):1--11.

\bibitem[{Inoue et~al.(2023{\natexlab{a}})Inoue, Kikuchi, Simo-Serra, Otani, and Yamaguchi}]{layoutdm}
Naoto Inoue, Kotaro Kikuchi, Edgar Simo-Serra, Mayu Otani, and Kota Yamaguchi. 2023{\natexlab{a}}.
\newblock Layoutdm: Discrete diffusion model for controllable layout generation.
\newblock In \emph{CVPR}, pages 10167--10176.

\bibitem[{Inoue et~al.(2023{\natexlab{b}})Inoue, Kikuchi, Simo-Serra, Otani, and Yamaguchi}]{flexdm}
Naoto Inoue, Kotaro Kikuchi, Edgar Simo-Serra, Mayu Otani, and Kota Yamaguchi. 2023{\natexlab{b}}.
\newblock Towards flexible multi-modal document models.
\newblock In \emph{CVPR}, pages 14287--14296.

\bibitem[{Inoue et~al.(2024)Inoue, Masui, Shimoda, and Yamaguchi}]{opencole}
Naoto Inoue, Kento Masui, Wataru Shimoda, and Kota Yamaguchi. 2024.
\newblock Opencole: Towards reproducible automatic graphic design generation.
\newblock In \emph{CVPR Workshops}, pages 8131--8135.

\bibitem[{Islam et~al.(2024)Islam, Ali, and Parvez}]{islam2024mapcoder}
Md~Ashraful Islam, Mohammed~Eunus Ali, and Md~Rizwan Parvez. 2024.
\newblock Mapcoder: Multi-agent code generation for competitive problem solving.
\newblock In \emph{ACL: Long Papers}, pages 4912--4944.

\bibitem[{Iwai et~al.(2024)Iwai, Osanai, Kitada, and Omachi}]{iwai2024layout}
Shoma Iwai, Atsuki Osanai, Shunsuke Kitada, and Shinichiro Omachi. 2024.
\newblock Layout-corrector: Alleviating layout sticking phenomenon in discrete diffusion model.
\newblock In \emph{ECCV}, pages 92--110. Springer.

\bibitem[{Jia et~al.(2023)Jia, Li, Yuan, Liu, Shen, Chen, Chen, Zheng, Chen, Li et~al.}]{cole}
Peidong Jia, Chenxuan Li, Yuhui Yuan, Zeyu Liu, Yichao Shen, Bohan Chen, Xingru Chen, Yinglin Zheng, Dong Chen, Ji~Li, and 1 others. 2023.
\newblock Cole: A hierarchical generation framework for multi-layered and editable graphic design.
\newblock \emph{arXiv preprint arXiv:2311.16974}.

\bibitem[{Katai(2024)}]{metricsurvey1}
Robert Katai. 2024.
\newblock \href {https://www.omniconvert.com/blog/5-powerful-banner-ad-ab-testing-ideas-try/} {5 powerful banner ad a/b testing ideas you should try}.
\newblock Accessed: 2025-02-28.

\bibitem[{Kikuchi et~al.(2024)Kikuchi, Inoue, Otani, Simo-Serra, and Yamaguchi}]{markupdm}
Kotaro Kikuchi, Naoto Inoue, Mayu Otani, Edgar Simo-Serra, and Kota Yamaguchi. 2024.
\newblock Multimodal markup document models for graphic design completion.
\newblock \emph{arXiv preprint arXiv:2409.19051}.

\bibitem[{King(2020)}]{metricsurvey3}
Cassandra King. 2020.
\newblock \href {https://www.superside.com/blog/banner-ad-design} {Essential practices for high-converting banner ad designs}.
\newblock Accessed: 2025-02-28.

\bibitem[{Koh et~al.(2024)Koh, Lo, Jang, Duvvur, Lim, Huang, Neubig, Zhou, Salakhutdinov, and Fried}]{web1}
Jing~Yu Koh, Robert Lo, Lawrence Jang, Vikram Duvvur, Ming Lim, Po-Yu Huang, Graham Neubig, Shuyan Zhou, Russ Salakhutdinov, and Daniel Fried. 2024.
\newblock Visualwebarena: Evaluating multimodal agents on realistic visual web tasks.
\newblock In \emph{ACL: Long Papers)}, pages 881--905.

\bibitem[{Landa(2016)}]{metricsurvey4}
Robin Landa. 2016.
\newblock \emph{Advertising by design: generating and designing creative ideas across media}.
\newblock John Wiley \& Sons.

\bibitem[{Lee et~al.(2020)Lee, Jiang, Essa, Le, Gong, Yang, and Yang}]{lee2020neuraleccv}
Hsin-Ying Lee, Lu~Jiang, Irfan Essa, Phuong~B Le, Haifeng Gong, Ming-Hsuan Yang, and Weilong Yang. 2020.
\newblock Neural design network: Graphic layout generation with constraints.
\newblock In \emph{ECCV}, pages 491--506. Springer.

\bibitem[{Li et~al.(2024)Li, Yan, Pan, Luo, Ji, Ding, Xu, Liu, Dong, Lin et~al.}]{li2024mmedagent}
Binxu Li, Tiankai Yan, Yuanting Pan, Jie Luo, Ruiyang Ji, Jiayuan Ding, Zhe Xu, Shilong Liu, Haoyu Dong, Zihao Lin, and 1 others. 2024.
\newblock Mmedagent: Learning to use medical tools with multi-modal agent.
\newblock In \emph{Findings of EMNLP}, pages 8745--8760.

\bibitem[{Li and Yang(2023)}]{li2023smartbanner}
Guandong Li and Xian Yang. 2023.
\newblock Smartbanner: intelligent banner design framework that strikes a balance between creative freedom and design rules.
\newblock \emph{Multimedia Tools and Applications}, 82(12):18653--18667.

\bibitem[{Li et~al.(2023)Li, Li, Savarese, and Hoi}]{mllm4}
Junnan Li, Dongxu Li, Silvio Savarese, and Steven Hoi. 2023.
\newblock Blip-2: Bootstrapping language-image pre-training with frozen image encoders and large language models.
\newblock In \emph{ICML}, pages 19730--19742. PMLR.

\bibitem[{Lin et~al.(2023{\natexlab{a}})Lin, Guo, Sun, Yang, Lou, and Zhang}]{layoutprompter}
Jiawei Lin, Jiaqi Guo, Shizhao Sun, Zijiang Yang, Jian-Guang Lou, and Dongmei Zhang. 2023{\natexlab{a}}.
\newblock Layoutprompter: awaken the design ability of large language models.
\newblock In \emph{NeurIPS}, volume~36, pages 43852--43879.

\bibitem[{Lin et~al.(2025)Lin, Sun, Huang, Liu, Li, and Bian}]{lin2024elements}
Jiawei Lin, Shizhao Sun, Danqing Huang, Ting Liu, Ji~Li, and Jiang Bian. 2025.
\newblock From elements to design: A layered approach for automatic graphic design composition.
\newblock In \emph{CVPR}.

\bibitem[{Lin et~al.(2024{\natexlab{a}})Lin, Huang, Zhao, Zhan, and Lin}]{lin2024spot}
Jieru Lin, Danqing Huang, Tiejun Zhao, Dechen Zhan, and Chin-Yew Lin. 2024{\natexlab{a}}.
\newblock Spot the error: Non-autoregressive graphic layout generation with wireframe locator.
\newblock In \emph{AAAI}, volume~38, pages 3413--3421.

\bibitem[{Lin et~al.(2023{\natexlab{b}})Lin, Zhou, Ma, Gao, Fei, Chen, Yu, and Ge}]{lin2023autoposter}
Jinpeng Lin, Min Zhou, Ye~Ma, Yifan Gao, Chenxi Fei, Yangjian Chen, Zhang Yu, and Tiezheng Ge. 2023{\natexlab{b}}.
\newblock Autoposter: A highly automatic and content-aware design system for advertising poster generation.
\newblock In \emph{ACM MM}, pages 1250--1260.

\bibitem[{Lin et~al.(2024{\natexlab{b}})Lin, Hua, Li, Chang, Fan, Ji, Hua, Jin, Luo, and Zhang}]{lin2024battleagent}
Shuhang Lin, Wenyue Hua, Lingyao Li, Che-Jui Chang, Lizhou Fan, Jianchao Ji, Hang Hua, Mingyu Jin, Jiebo Luo, and Yongfeng Zhang. 2024{\natexlab{b}}.
\newblock Battleagent: Multi-modal dynamic emulation on historical battles to complement historical analysis.
\newblock In \emph{EMNLP: System Demonstrations}, pages 172--181.

\bibitem[{Liu et~al.(2023)Liu, Li, Wu, and Lee}]{mllm1}
Haotian Liu, Chunyuan Li, Qingyang Wu, and Yong~Jae Lee. 2023.
\newblock Visual instruction tuning.
\newblock In \emph{NeurIPS}, volume~36, pages 34892--34916.

\bibitem[{Liu et~al.(2024)Liu, Liang, Liang, Luo, Li, Huang, and Yuan}]{glyphbyt5}
Zeyu Liu, Weicong Liang, Zhanhao Liang, Chong Luo, Ji~Li, Gao Huang, and Yuhui Yuan. 2024.
\newblock Glyph-byt5: A customized text encoder for accurate visual text rendering.
\newblock In \emph{ECCV}, pages 361--377. Springer.

\bibitem[{Lohtia et~al.(2003)Lohtia, Donthu, and Hershberger}]{metricsurvey5}
Ritu Lohtia, Naveen Donthu, and Edmund~K Hershberger. 2003.
\newblock The impact of content and design elements on banner advertising click-through rates.
\newblock \emph{Journal of advertising Research}, 43(4):410--418.

\bibitem[{Lu et~al.(2024)Lu, Lu, Lange, Foerster, Clune, and Ha}]{lu2024ai}
Chris Lu, Cong Lu, Robert~Tjarko Lange, Jakob Foerster, Jeff Clune, and David Ha. 2024.
\newblock The ai scientist: Towards fully automated open-ended scientific discovery.
\newblock \emph{arXiv preprint arXiv:2408.06292}.

\bibitem[{Madaan et~al.(2023)Madaan, Tandon, Gupta, Hallinan, Gao, Wiegreffe, Alon, Dziri, Prabhumoye, Yang et~al.}]{selfrefine}
Aman Madaan, Niket Tandon, Prakhar Gupta, Skyler Hallinan, Luyu Gao, Sarah Wiegreffe, Uri Alon, Nouha Dziri, Shrimai Prabhumoye, Yiming Yang, and 1 others. 2023.
\newblock Self-refine: Iterative refinement with self-feedback.
\newblock In \emph{NeurIPS}, volume~36, pages 46534--46594.

\bibitem[{Maheshwari et~al.(2019)Maheshwari, Bansal, Dwivedi, Kumar, Manerikar, and Srinivasan}]{maheshwari2019exemplar}
Paridhi Maheshwari, Nitish Bansal, Surya Dwivedi, Rohan Kumar, Pranav Manerikar, and Balaji~Vasan Srinivasan. 2019.
\newblock Exemplar based experience transfer.
\newblock In \emph{IUI}, pages 673--680.

\bibitem[{Malhotra(2024)}]{metricsurvey2}
Smridhi Malhotra. 2024.
\newblock \href {https://duck.design/practices-for-high-converting-banner-ad-designs/} {Essential practices for high-converting banner ad designs}.
\newblock Accessed: 2025-02-28.

\bibitem[{ManyPixels(2025)}]{designerprice}
ManyPixels. 2025.
\newblock \href {https://www.manypixels.co/blog/get-a-designer/graphic-design-price} {The complete graphic design pricing guide (2025 update)}.
\newblock Accessed: 2025-02-28.

\bibitem[{OpenAI(2024)}]{gpt4o}
OpenAI. 2024.
\newblock \href {https://openai.com/index/hello-gpt-4o/} {Hello gpt-4o}.

\bibitem[{O’Donovan et~al.(2014)O’Donovan, Agarwala, and Hertzmann}]{o2014learning}
Peter O’Donovan, Aseem Agarwala, and Aaron Hertzmann. 2014.
\newblock Learning layouts for single-pagegraphic designs.
\newblock \emph{TVCG}, 20(8):1200--1213.

\bibitem[{Park et~al.(2024)Park, Bae, Shin, and Jeon}]{park2024kinetic}
Seonmi Park, Inhwan Bae, Seunghyun Shin, and Hae-Gon Jeon. 2024.
\newblock Kinetic typography diffusion model.
\newblock In \emph{ECCV}, pages 166--185. Springer.

\bibitem[{Qian et~al.(2024)Qian, Liu, Liu, Chen, Dang, Li, Yang, Chen, Su, Cong et~al.}]{qian2024chatdev}
Chen Qian, Wei Liu, Hongzhang Liu, Nuo Chen, Yufan Dang, Jiahao Li, Cheng Yang, Weize Chen, Yusheng Su, Xin Cong, and 1 others. 2024.
\newblock Chatdev: Communicative agents for software development.
\newblock In \emph{ACL: Long Papers}, pages 15174--15186.

\bibitem[{Robinson et~al.(2007)Robinson, Wysocka, and Hand}]{metricsurvey6}
Helen Robinson, Anna Wysocka, and Chris Hand. 2007.
\newblock Internet advertising effectiveness: the effect of design on click-through rates for banner ads.
\newblock \emph{International journal of advertising}, 26(4):527--541.

\bibitem[{Rombach et~al.(2022)Rombach, Blattmann, Lorenz, Esser, and Ommer}]{stablediffusion}
Robin Rombach, Andreas Blattmann, Dominik Lorenz, Patrick Esser, and Bj{\"o}rn Ommer. 2022.
\newblock High-resolution image synthesis with latent diffusion models.
\newblock In \emph{CVPR}, pages 10684--10695.

\bibitem[{Shabani et~al.(2024)Shabani, Wang, Liu, Zhao, Yang, and Furukawa}]{vlc}
Mohammad~Amin Shabani, Zhaowen Wang, Difan Liu, Nanxuan Zhao, Jimei Yang, and Yasutaka Furukawa. 2024.
\newblock Visual layout composer: Image-vector dual diffusion model for design layout generation.
\newblock In \emph{CVPR}, pages 9222--9231.

\bibitem[{Shimoda et~al.(2024)Shimoda, Inoue, Haraguchi, Mitani, Uchida, and Yamaguchi}]{typer}
Wataru Shimoda, Naoto Inoue, Daichi Haraguchi, Hayato Mitani, Seichi Uchida, and Kota Yamaguchi. 2024.
\newblock Type-r: Automatically retouching typos for text-to-image generation.
\newblock \emph{arXiv preprint arXiv:2411.18159}.

\bibitem[{Tuo et~al.(2024)Tuo, Xiang, He, Geng, and Xie}]{tuoanytext}
Yuxiang Tuo, Wangmeng Xiang, Jun-Yan He, Yifeng Geng, and Xuansong Xie. 2024.
\newblock Anytext: Multilingual visual text generation and editing.
\newblock In \emph{ICLR}.

\bibitem[{Vaddamanu et~al.(2022)Vaddamanu, Aggarwal, Guda, Srinivasan, and Chhaya}]{vaddamanu2022harmonized}
Praneetha Vaddamanu, Vinay Aggarwal, Bhanu Prakash~Reddy Guda, Balaji~Vasan Srinivasan, and Niyati Chhaya. 2022.
\newblock Harmonized banner creation from multimodal design assets.
\newblock In \emph{CHI Extended Abstracts}, pages 1--7.

\bibitem[{Vempati et~al.(2020)Vempati, Malayil, Sruthi, and Sandeep}]{vempati2020enabling}
Sreekanth Vempati, Korah~T Malayil, V~Sruthi, and R~Sandeep. 2020.
\newblock Enabling hyper-personalisation: Automated ad creative generation and ranking for fashion e-commerce.
\newblock In \emph{Fashion Recommender Systems}, pages 25--48. Springer.

\bibitem[{Vinker et~al.(2024)Vinker, Shaham, Zheng, Zhao, Fan, and Torralba}]{sketchagent}
Yael Vinker, Tamar~Rott Shaham, Kristine Zheng, Alex Zhao, Judith~E Fan, and Antonio Torralba. 2024.
\newblock Sketchagent: Language-driven sequential sketch generation.
\newblock \emph{arXiv preprint arXiv:2411.17673}.

\bibitem[{Wang et~al.(2024)Wang, Ge, Chen, Zhou, Wang, Cheng, and Yuan}]{prompt2poster}
Shaodong Wang, Yunyang Ge, Liuhan Chen, Haiyang Zhou, Qian Wang, Xinhua Cheng, and Li~Yuan. 2024.
\newblock Prompt2poster: Automatically artistic chinese poster creation from prompt only.
\newblock In \emph{ACM MM}, pages 10716--10724.

\bibitem[{Wang et~al.(2022)Wang, Liu, Zhong, Zhou, Ge, Lian, and Jiang}]{creagan}
Shiyao Wang, Qi~Liu, Yicheng Zhong, Zhilong Zhou, Tiezheng Ge, Defu Lian, and Yuning Jiang. 2022.
\newblock Creagan: An automatic creative generation framework for display advertising.
\newblock In \emph{ACM MM}, pages 7261--7269.

\bibitem[{Wang et~al.(2025)Wang, Li, Li, and Liu}]{genartist}
Zhenyu Wang, Aoxue Li, Zhenguo Li, and Xihui Liu. 2025.
\newblock Genartist: Multimodal llm as an agent for unified image generation and editing.
\newblock In \emph{NeurIPS}, volume~37, pages 128374--128395.

\bibitem[{Weng et~al.(2024)Weng, Huang, Qiao, Hu, Lin, Zhang, and Chen}]{weng2024desigen}
Haohan Weng, Danqing Huang, Yu~Qiao, Zheng Hu, Chin-Yew Lin, Tong Zhang, and CL~Chen. 2024.
\newblock Desigen: A pipeline for controllable design template generation.
\newblock In \emph{CVPR}, pages 12721--12732.

\bibitem[{Wu et~al.(2024)Wu, Xu, and Wang}]{wu2024transagents}
Minghao Wu, Jiahao Xu, and Longyue Wang. 2024.
\newblock Transagents: Build your translation company with language agents.
\newblock In \emph{EMNLP: System Demonstrations}, pages 131--141.

\bibitem[{Xie et~al.(2024)Xie, Chen, Zhang, Wan, and Li}]{mmagent_survey2}
Junlin Xie, Zhihong Chen, Ruifei Zhang, Xiang Wan, and Guanbin Li. 2024.
\newblock Large multimodal agents: A survey.
\newblock \emph{arXiv preprint arXiv:2402.15116}.

\bibitem[{Xu et~al.(2024)Xu, Hu, Zhou, Ren, Dong, Keutzer, Ng, and Feng}]{xu2024magic}
Lin Xu, Zhiyuan Hu, Daquan Zhou, Hongyu Ren, Zhen Dong, Kurt Keutzer, See~Kiong Ng, and Jiashi Feng. 2024.
\newblock Magic: Investigation of large language model powered multi-agent in cognition, adaptability, rationality and collaboration.
\newblock In \emph{EMNLP}, pages 7315--7332.

\bibitem[{Yang et~al.(2024{\natexlab{a}})Yang, Yuan, Yang, Xu, Yuan, and Zeng}]{yang2024new}
Hao Yang, Jianxin Yuan, Shuai Yang, Linhe Xu, Shuo Yuan, and Yifan Zeng. 2024{\natexlab{a}}.
\newblock A new creative generation pipeline for click-through rate with stable diffusion model.
\newblock In \emph{WWW}, pages 180--189.

\bibitem[{Yang et~al.(2016)Yang, Mei, Xu, Rui, and Li}]{yang2016automatic}
Xuyong Yang, Tao Mei, Ying-Qing Xu, Yong Rui, and Shipeng Li. 2016.
\newblock Automatic generation of visual-textual presentation layout.
\newblock \emph{TOMM}, 12(2):1--22.

\bibitem[{Yang et~al.(2024{\natexlab{b}})Yang, Wang, Li, Lin, Lin, Liu, and Wang}]{idea2img}
Zhengyuan Yang, Jianfeng Wang, Linjie Li, Kevin Lin, Chung-Ching Lin, Zicheng Liu, and Lijuan Wang. 2024{\natexlab{b}}.
\newblock Idea2img: Iterative self-refinement with gpt-4v (ision) for automatic image design and generation.
\newblock In \emph{ECCV}.

\bibitem[{Yao et~al.(2023)Yao, Zhao, Yu, Du, Shafran, Narasimhan, and Cao}]{react}
Shunyu Yao, Jeffrey Zhao, Dian Yu, Nan Du, Izhak Shafran, Karthik Narasimhan, and Yuan Cao. 2023.
\newblock React: Synergizing reasoning and acting in language models.
\newblock In \emph{ICLR}.

\bibitem[{Yun et~al.(2025)Yun, Wang, Shimose, Choo, and Takamatsu}]{yun2025designlab}
Jooyeol Yun, Heng Wang, Yotaro Shimose, Jaegul Choo, and Shingo Takamatsu. 2025.
\newblock Designlab: Designing slides through iterative detection and correction.
\newblock \emph{arXiv preprint arXiv:2507.17202}.

\bibitem[{Zhang et~al.(2024)Zhang, Xu, Zhang, Liu, Hooi, and Deng}]{zhang2024exploring}
Jintian Zhang, Xin Xu, Ningyu Zhang, Ruibo Liu, Bryan Hooi, and Shumin Deng. 2024.
\newblock Exploring collaboration mechanisms for llm agents: A social psychology view.
\newblock In \emph{ACL: Long Papers}, pages 14544--14607.

\bibitem[{Zhang et~al.(2023)Zhang, Rao, and Agrawala}]{controlnet}
Lvmin Zhang, Anyi Rao, and Maneesh Agrawala. 2023.
\newblock Adding conditional control to text-to-image diffusion models.
\newblock In \emph{ICCV}, pages 3836--3847.

\bibitem[{Zhang et~al.(2017)Zhang, Hu, Ren, Yang, Xu, and Hua}]{zhang2017layout}
Yunke Zhang, Kangkang Hu, Peiran Ren, Changyuan Yang, Weiwei Xu, and Xian-Sheng Hua. 2017.
\newblock Layout style modeling for automating banner design.
\newblock In \emph{ACM MM Workshops}, pages 451--459.

\bibitem[{Zhao et~al.(2024)Zhao, Tang, Wu, Lin, Wei, Liu, Tan, Huang, Xie et~al.}]{zhaoharmonizing}
Zhen Zhao, Jingqun Tang, Binghong Wu, Chunhui Lin, Shu Wei, Hao Liu, Xin Tan, Can Huang, Yuan Xie, and 1 others. 2024.
\newblock Harmonizing visual text comprehension and generation.
\newblock In \emph{NeurIPS}.

\bibitem[{Zheng et~al.(2025)Zheng, Guan, Kong, Zheng, Lin, Lu, He, Han, and Sun}]{pptagent}
Hao Zheng, Xinyan Guan, Hao Kong, Jia Zheng, Hongyu Lin, Yaojie Lu, Ben He, Xianpei Han, and Le~Sun. 2025.
\newblock Pptagent: Generating and evaluating presentations beyond text-to-slides.
\newblock \emph{arXiv preprint arXiv:2501.03936}.

\bibitem[{Zheng et~al.(2019)Zheng, Qiao, Cao, and Lau}]{zheng2019content}
Xinru Zheng, Xiaotian Qiao, Ying Cao, and Rynson~WH Lau. 2019.
\newblock Content-aware generative modeling of graphic design layouts.
\newblock \emph{TOG}, 38(4):1--15.

\bibitem[{Zhu et~al.(2024)Zhu, Chen, Shen, Li, and Elhoseiny}]{mllm2}
Deyao Zhu, Jun Chen, Xiaoqian Shen, Xiang Li, and Mohamed Elhoseiny. 2024.
\newblock Minigpt-4: Enhancing vision-language understanding with advanced large language models.
\newblock In \emph{ICLR}.

\end{thebibliography}

\clearpage
\appendix

\section{Appendix}
\label{sec:appendix}
In this appendix, we present more details on our experimental setup (Sec.~\ref{sec:setup}), human study setup (Sec.~\ref{sec:hs}), text-free background generation (Sec.~\ref{sec:dg-a}), and foreground variation generation (Sec.~\ref{sec:dg-b}). We also show detailed statistics results on the refinement human study (Sec.~\ref{sec:ra-a}) and general graphic design evaluation (Sec.~\ref{sec:ra-b}). The detailed prompt for text-to-image generation models in Fig.~\ref{fig:visual_compare} (a) of our main paper is presented in Sec.~\ref{sec:dp}. The agent profiles are displayed in Sec.~\ref{sec:agent_profile} Lastly, we showcase more results from our BannerAgency in Sec.~\ref{sec:moreresults}.
\subsection{More Details on Experimental Setup}
\label{sec:setup}

\paragraph{Cost analysis.} We present a detailed cost breakdown in USD for generating banner advertisements of size $300\times250$ pixels in Tab.~\ref{tab:cost_breakdown}. Our BannerAgency upscales images to $1024\times864$ pixels to maintain visual quality before rendering. Costs are calculated based on Claude 3.5 Sonnet API pricing (\$3/MTok for input, \$15/MTok for output) and FLUX-1 [schnell] image generation (\$0.0027/MP). We compare both Figma and SVG implementation approaches across three scenarios: single image generation, generation with four-rounds of iterative refinement, and multi-variation generation with four design alternatives as detailed in Sec.~\ref{sec:dg-b}. The dramatic difference between the two implementations stems from fundamental implementation constraints: while SVG can be generated directly as code that LLMs frequently encounter in training data, Figma plugin JavaScript is uncommon in LLM training corpora. To ensure reliable execution, our Figma implementation requires providing templates where the Developer agent only needs to call predefined functions, significantly increasing token usage. Notably, both implementation approaches offer extraordinary cost-effectiveness compared to hiring professional designers, who typically charge \$50-150 per hour~\cite{designerprice}. Even our most expensive scenario (Figma with feedback at \$1.213) represents less than 2.5\% of traditional design costs, offering transformative economics for advertising campaigns requiring multiple banner variations.

\begin{table}[htbp]
\centering
\LARGE
\resizebox{\columnwidth}{!}{
\begin{tabular}{llcccccc}
\hline
\textbf{Implementation} & \textbf{Scenario} & \textbf{Prompt} & \textbf{Completion} & \textbf{LLM Cost} & \textbf{Image Cost} & \textbf{Total Cost} & \textbf{Cost/Image} \\
 & & \textbf{Tokens} & \textbf{Tokens} & \textbf{(\$)} & \textbf{(\$)} & \textbf{(\$)} & \textbf{(\$)} \\
\hline
\multirow{2}{*}{Single image} & Figma & 78,266 & 5,470 & 0.317 & 0.002 & 0.319 & 0.319 \\
 & SVG & 15,921 & 2,672 & 0.088 & 0.002 & 0.090 & 0.090 \\
\hline
\multirow{2}{*}{With refinement} & Figma & 284,677 & 23,772 & 1.211 & 0.002 & 1.213 & 1.213 \\
 & SVG & 54,362 & 12,663 & 0.353 & 0.002 & 0.355 & 0.355 \\
\hline
\multirow{2}{*}{Four variations} & Figma & 261,150 & 15,213 & 1.012 & 0.008 & 1.020 & 0.255 \\
 & SVG & 24,865 & 6,006 & 0.165 & 0.008 & 0.173 & 0.043 \\
\hline
\end{tabular}}
\caption{BannerAgency System Cost Breakdown}
\label{tab:cost_breakdown}
\end{table}

\begin{figure*}[!htb]
\centering     
\includegraphics[width=\linewidth]{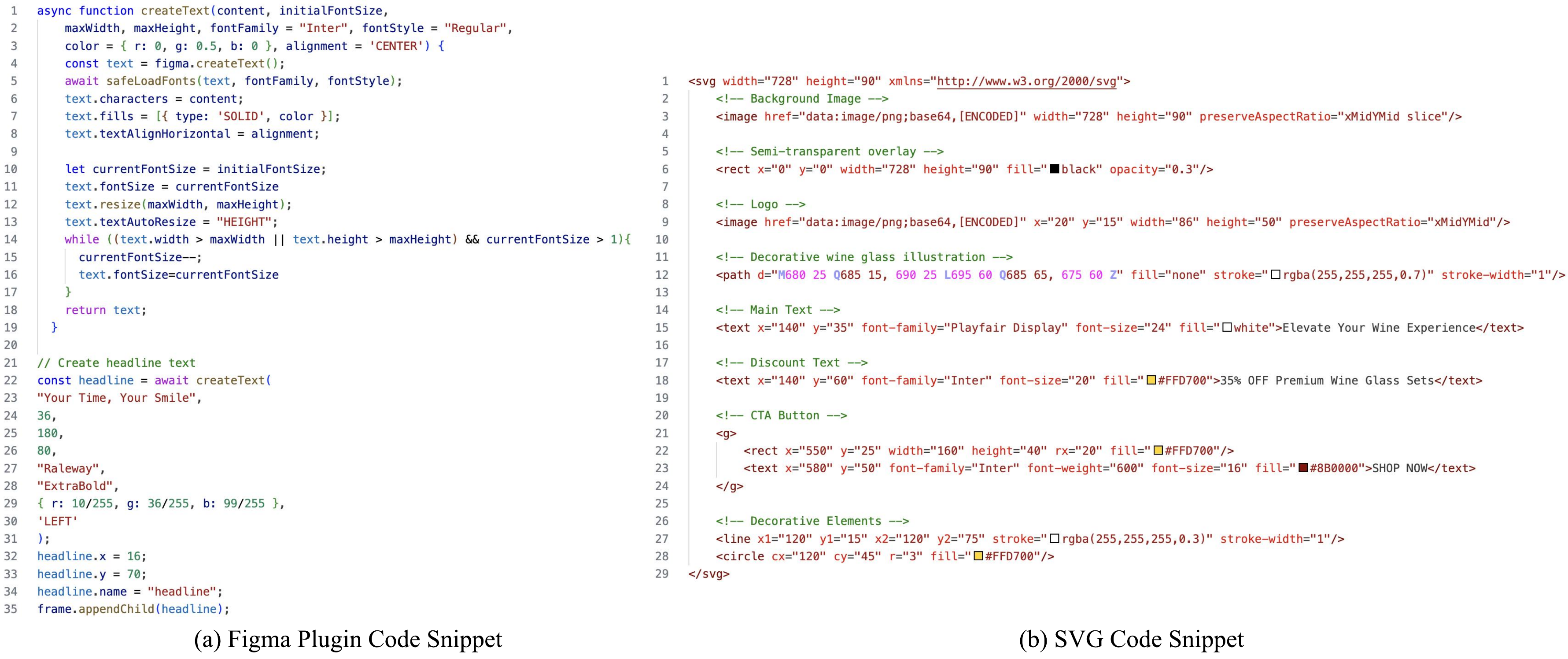}
\caption{Figma vs. SVG code format.}
\label{fig:code_format}
\end{figure*}

\paragraph{Figma vs. SVG format}
Both implementation approaches have their merits, depending on the user's preferences and the design tools they are comfortable with. The SVG code generation offers broad compatibility and flexibility, while the Figma plugin code generation provides a more integrated and automated experience within the Figma ecosystem. The code examples are provided in Fig.~\ref{fig:code_format}.

\paragraph{Agent profile definition.} Please refer to Sec.~\ref{sec:agent_profile} for each agent's profile definition.
\paragraph{Agent tool definition.}
We document the tool list in Tab.~\ref{tab:developer_tools}. The Background Designer agent creates visual backdrops for banner advertisements that align with brand identity and campaign objectives. It leverages existing assets when available and employs a refinement process to ensure text-free backgrounds through detection and regeneration cycles. The system handles dimensional requirements by adapting to supported aspect ratios and resizing outputs to match target specifications. The Developer Agent transforms design specifications into editable implementations through a systematic process. It loads and customizes Figma plugin templates, manages asset references, compiles implementation code, and handles rendering by interfacing with Figma's API. Throughout this workflow, comprehensive archiving preserves complete implementations alongside their assets, enabling advertisers to receive fully editable component-based designs rather than static images.

\begin{table*}[!htb]

\centering
\small
\begin{tabular}{p{0.25\linewidth}|p{0.7\linewidth}}
\toprule
\textbf{Tool Name} & \textbf{Description} \\
\midrule
\texttt{update\_image\_list} & Updates the imageList constant in a Figma plugin's UI HTML file with a provided list of image filenames, enabling dynamic referencing of assets within the plugin interface. \\
\midrule
\texttt{read\_plugin\_template} & Reads and validates a Figma plugin code template file, ensuring it's a proper JavaScript file containing essential Figma plugin components before using it as a foundation for customization. \\
\midrule
\texttt{save\_plugin\_code} & Saves the finalized JavaScript code to the code.js file in the specified Figma plugin working directory after validating that the directory contains a manifest.json file. \\
\midrule
\texttt{render\_and\_save\_image} & Executes the Figma plugin by running an AppleScript, retrieves the latest node ID, and uses the Figma API to export and save the rendered banner design as a PNG image at the specified path. \\
\midrule
\texttt{backup\_plugin\_folder} & Creates a backup copy of the entire plugin folder with all assets (including background image, logo, and rendered output) to preserve the complete state of the implementation for future reference or versioning. \\
\midrule
\texttt{create\_unique\_image\_name} & Generates a unique filename by combining a timestamp with the banner name to ensure distinct identification of each banner design output (e.g., "20240328\_154227\_kids\_dentist\_clinic\_banner"). \\
\midrule
\texttt{T2I} & Generates banner backgrounds from textual descriptions by interfacing with a text-to-image model, adapting dimensions to supported aspect ratios, and resizing outputs to match target banner dimensions. Includes negative prompting to avoid text generation in backgrounds. \\
\midrule
\texttt{FindImagePath} & Analyzes user input to locate existing image files that could serve as backgrounds, identifies valid image paths with common extensions (.jpg, .png, etc.), filters out logo images, verifies file existence, and resizes found images to match target banner dimensions. \\
\midrule
\texttt{TextChecker} & Evaluates generated or existing background images to detect the presence of any text elements, using a multimodal LLM to analyze the visual content and determine if the background is suitable or requires regeneration. \\
\bottomrule
\end{tabular}
\caption{List of tool definitions.}
\label{tab:developer_tools}
\end{table*}
\paragraph{Metrics rating criteria.}
We present the detailed rating criteria for each metric in Tab.~\ref{tab:banner_criteria}.
\begin{table*}[htbp]
\centering
\small
\renewcommand{\arraystretch}{1.2}
\begin{tabular}{p{0.05\textwidth}p{0.15\textwidth}p{0.2\textwidth}p{0.5\textwidth}}
\hline
\textbf{Abbr.} & \textbf{Metric Name} & \textbf{Definition} & \textbf{Rating Criteria} \\
\hline
TAA & Target Audience Alignment & Measures how well the generated banner ad aligns with the given request, including the theme, target audience, and primary purpose. & 
\begin{minipage}[t]{0.5\textwidth}
\textbf{5} – Perfectly aligns with the request (theme, audience, purpose are all clearly reflected).

\textbf{4} – Mostly aligns, but minor details could be improved.

\textbf{3} – Somewhat aligns, but key elements are missing or unclear.

\textbf{2} – Barely aligns, with major missing or incorrect elements.

\textbf{1} – Does not align with the request at all.
\end{minipage} \\
\hline
LPS & Logo Placement Score & Evaluates whether the logo is well-integrated into the design in terms of visibility, size, and positioning. & 
\begin{minipage}[t]{0.5\textwidth}
\textbf{5} – Logo is well-placed, clearly visible, proportionate, and blends seamlessly.

\textbf{4} – Logo is well-placed but could be slightly improved (e.g., minor size or position adjustments).

\textbf{3} – Logo is visible but not ideally placed (e.g., too small, too large, or slightly obstructed).

\textbf{2} – Logo placement is poor (e.g., difficult to notice, awkward positioning).

\textbf{1} – Logo is either missing or completely misplaced.
\end{minipage} \\
\hline
AQS & Aesthetic Quality Score & Measures the visual appeal, including color harmony, layout balance, typography, and overall design quality. & 
\begin{minipage}[t]{0.5\textwidth}
\textbf{5} – Visually outstanding, professional design, well-balanced, with harmonious colors and readable text.

\textbf{4} – Well-designed, but small refinements could enhance it.

\textbf{3} – Acceptable but has notable design flaws (e.g., poor contrast, unbalanced elements).

\textbf{2} – Visually weak, with noticeable design mistakes.

\textbf{1} – Poor design, lacks professionalism or coherence.
\end{minipage} \\
\hline
CTAE & Call-to-Action Effectiveness & Evaluates whether the Call-to-Action (CTA) is clear, engaging, and visually emphasized. & 
\begin{minipage}[t]{0.5\textwidth}
\textbf{5} – CTA is clear, compelling, well-placed, and visually prominent.

\textbf{4} – CTA is effective but could be slightly improved (e.g., contrast, size).

\textbf{3} – CTA is present but lacks emphasis or clarity.

\textbf{2} – CTA is weak, hard to notice, or poorly worded.

\textbf{1} – No clear CTA is present.
\end{minipage} \\
\hline
CPYQ & Copywriting Quality & Evaluates the effectiveness of the headline, subheadline, and any other text in the banner ad, focusing on clarity, readability, persuasiveness, and grammatical correctness. & 
\begin{minipage}[t]{0.5\textwidth}
\textbf{5} – Copy is clear, engaging, grammatically correct, and persuasive, making the message effective.

\textbf{4} – Copy is well-written but could be slightly improved (e.g., minor word choice refinements).

\textbf{3} – Copy is somewhat effective but has issues in clarity, grammar, or persuasiveness.

\textbf{2} – Copy is weak, hard to read, contains noticeable grammatical mistakes, or lacks impact.

\textbf{1} – Copy is unclear, irrelevant, or difficult to read due to poor design or bad wording.
\end{minipage} \\
\hline
BIS & Brand Identity Score & Measures how well the banner ad visually and stylistically aligns with the brand's identity beyond just logo placement. This includes color consistency, typography, imagery, and overall brand feel. & 
\begin{minipage}[t]{0.5\textwidth}
\textbf{5} – Strong brand consistency; the banner design aligns well with the provided logo and conveys a recognizable brand identity.

\textbf{4} – Mostly aligns, but minor refinements could improve brand consistency.

\textbf{3} – Somewhat aligns, but noticeable inconsistencies exist (e.g., off-brand colors, incorrect typography).

\textbf{2} – Weak brand alignment, only the logo represents the brand while other design choices feel unrelated.

\textbf{1} – No brand identity is reflected; the banner appears generic or disconnected from the brand.\\

\end{minipage} \\
\hline
\end{tabular}
\caption{Banner design rating criteria with a 5-point scale for each metric.}
\label{tab:banner_criteria}
\end{table*}

\subsection{More Details on Human Study}
\label{sec:hs}
\begin{figure}[!htb]
\centering     
\includegraphics[width=\linewidth]{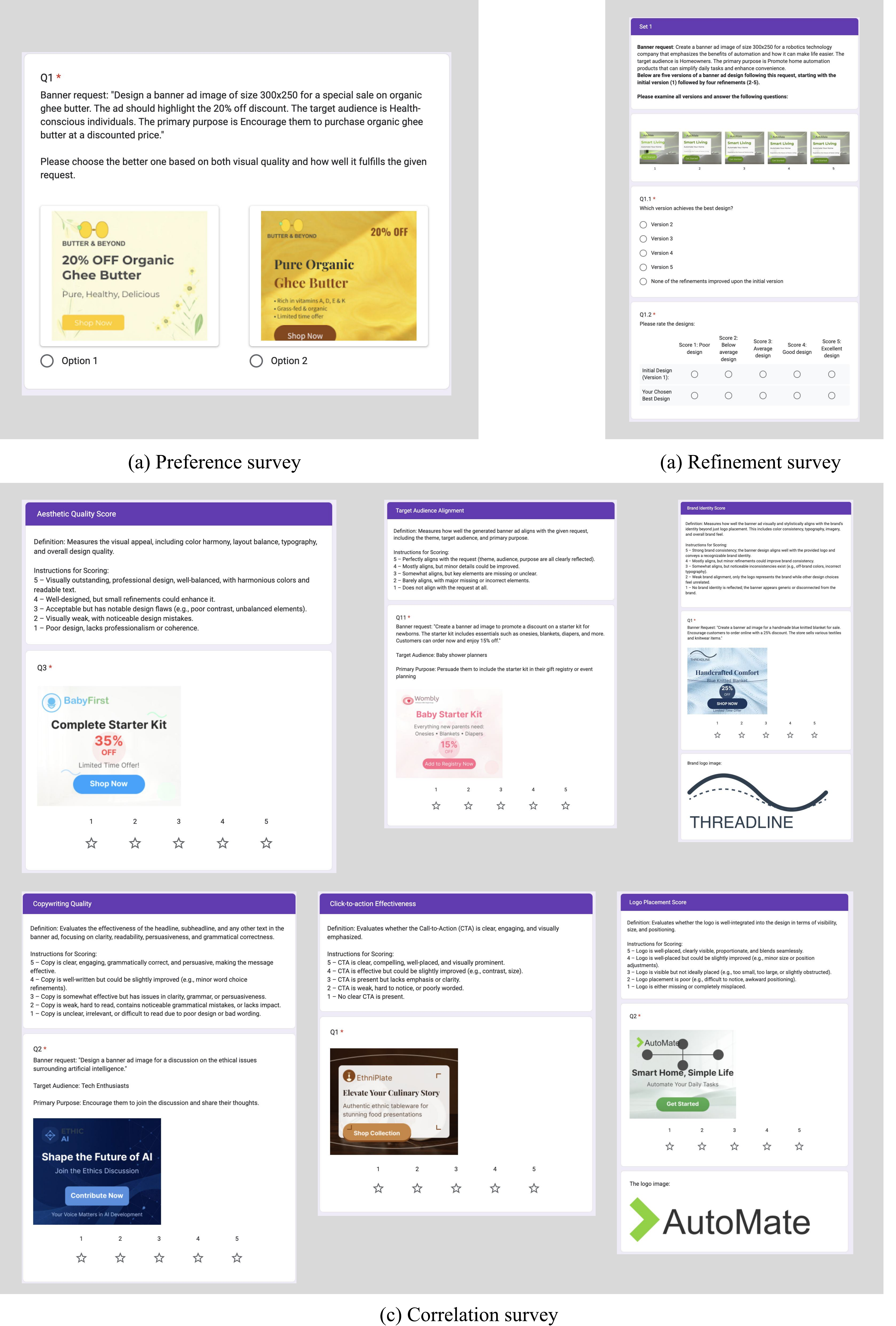}
\caption{The survey examples.}
\label{fig:all_survey}
\end{figure}

We present screenshots of the survey instructions in Fig.~\ref{fig:all_survey} to demonstrate how we structured each human evaluation.
\paragraph{Preference.}
The preference experiment evaluated user preferences across different variants of the banner ad generation system. It consisted of 20 banner ad requests, each implemented with 5 different system variants, resulting in 10 possible pairwise comparisons per request. Twenty participants completed the same set of comparisons, viewing pairs of banner ads generated from identical requests but using different system variants. For each pair, participants were asked to select which banner ad they preferred. The 20 banner requests were randomly selected to match the distribution of the full 400-request dataset, ensuring that the selected subset was representative of the overall performance characteristics of each variant. To avoid potential bias, the order of presentation for the banner pairs was randomized for each participant.
\paragraph{Refinement.}
The refinement experiment evaluated the effectiveness of an iterative refinement approach for banner ad image generation. It involved 20 sets of banner designs, with each set containing four iterations of refinement (modifying only foreground elements while keeping backgrounds consistent). Fifteen participants reviewed all 20 sets and performed two tasks: (1) selecting which iteration among the four achieved the best refinement performance, and (2) rating both the initial version and their chosen best version on a scale of 1 to 5. This experiment aimed to determine whether the refinement process consistently improved designs and at which iteration the optimal results were typically achieved.
\paragraph{Correlation.}
The correlation experiment assessed whether six proposed metrics for banner ad quality (TAA: Target Audience Alignment, LPS: Logo Placement Score, CTAE: CTA Effectiveness, CPYQ: Copywriting Quality, BIS: Brand Identity Score, and AQS: Aesthetic Quality Score) aligned with human perception. For each metric and each point on a 1-to-5 scale, five images previously graded by GPT-4o at that score level were selected, resulting in 25 images per metric and 150 images total. Ten participants graded these images, with the presentation order shuffled to prevent bias. Each participant evaluated all images across the six metrics to determine whether human evaluators would assign similar scores to those given by GPT-4o.


\subsection{More Details on Design Generation}
\paragraph{Details of Text-free Background Generation.}
\label{sec:dg-a}
Alg.~\ref{alg:banner_background} shows the detailed behavior of the text-free background generation. 
\begin{algorithm}[tb]
\small
    \caption{Text-free Background Design Agent Behavior}
    \label{alg:banner_background}
    \textbf{Input}: User requirements $R$, Logo characteristics $L$\\
    \textbf{Output}: Background image path $P$, Description $D$ (if generated) \\
    \textbf{Agent Tools}:\\
    \colorbox{gray!30}{
    \parbox{\linewidth}{
     $\textsc{FindImagePath}: R \mapsto (P, \text{bool})$ \quad \textit{Primary search tool}\\
    $\textsc{T2I}: D \mapsto P$ \quad \textit{Image generation tool}\\
    $\textsc{TextChecker}: P \mapsto \text{bool}$ \quad \textit{Verification tool}
    }}
    \begin{algorithmic}[1]
        \STATE $P, path\_exists \gets \textsc{FindImagePath}(R)$
        \IF{$path\_exists$}
            \STATE \textbf{return} $P$
        \ELSE
            \STATE $attempts \gets 0$
            \STATE $contains\_text \gets \texttt{TRUE}$
            \WHILE{$contains\_text$ AND $attempts < 5$}
                \STATE Analyze $R$ and $L$ to formulate background description $D$
                \STATE $P \gets \textsc{T2I}(D)$
                \STATE $contains\_text \gets \textsc{TextChecker}(P)$
                \STATE $attempts \gets attempts + 1$
                \IF{$contains\_text$}
                    \STATE Refine $D$ to eliminate text-producing elements
                \ENDIF
            \ENDWHILE
            \STATE \textbf{return} $P$, $D$
        \ENDIF
    \end{algorithmic}
\end{algorithm}

\paragraph{Memory-aware Design Variation.}
\label{sec:dg-b}

\begin{figure}[!htb]
\centering     
\includegraphics[width=\linewidth]{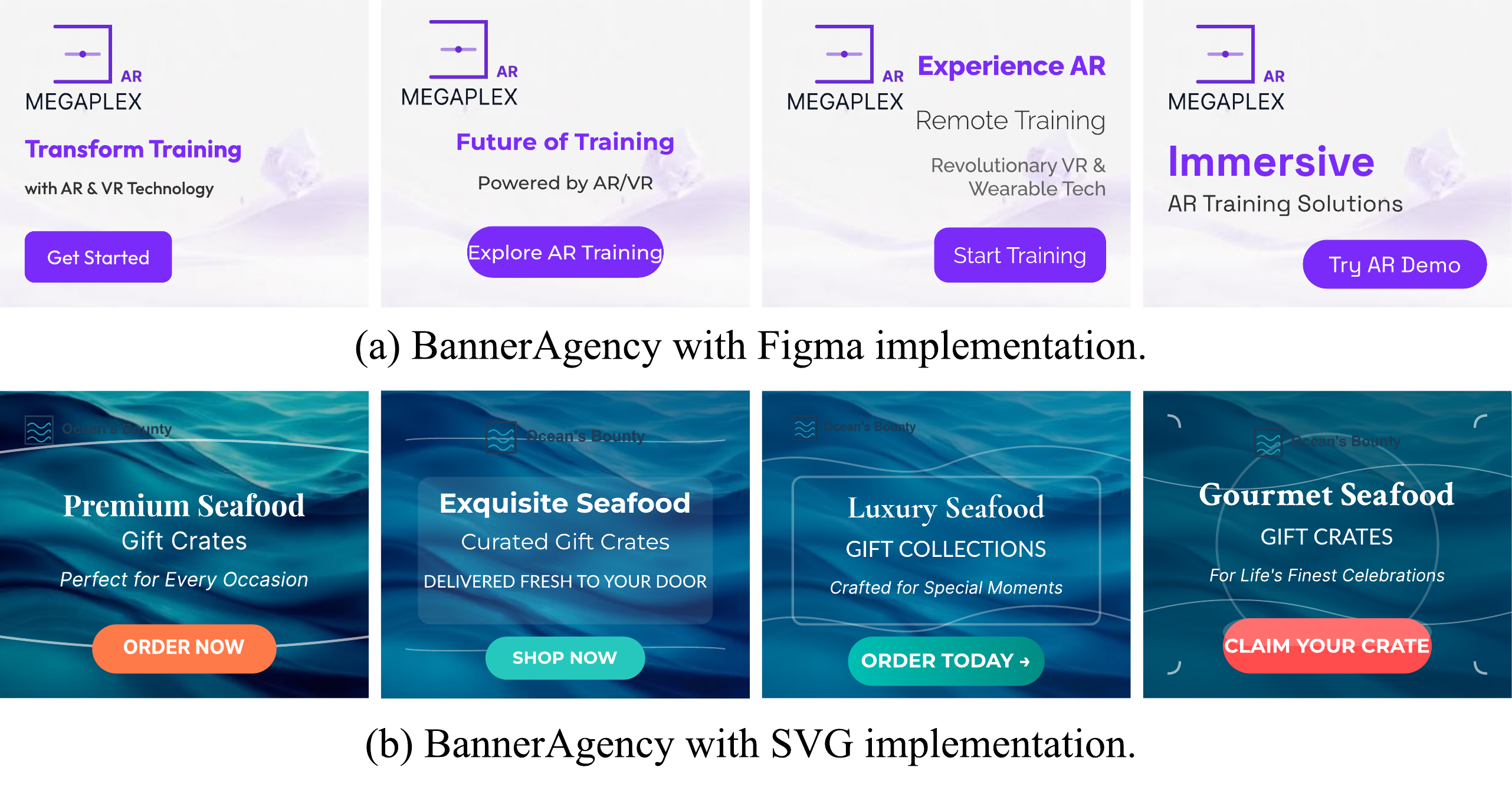}
\caption{Variation.}
\label{fig:diversity}
\end{figure}
\begin{figure*}[!htbp]
\centering     
\includegraphics[width=0.9\linewidth]{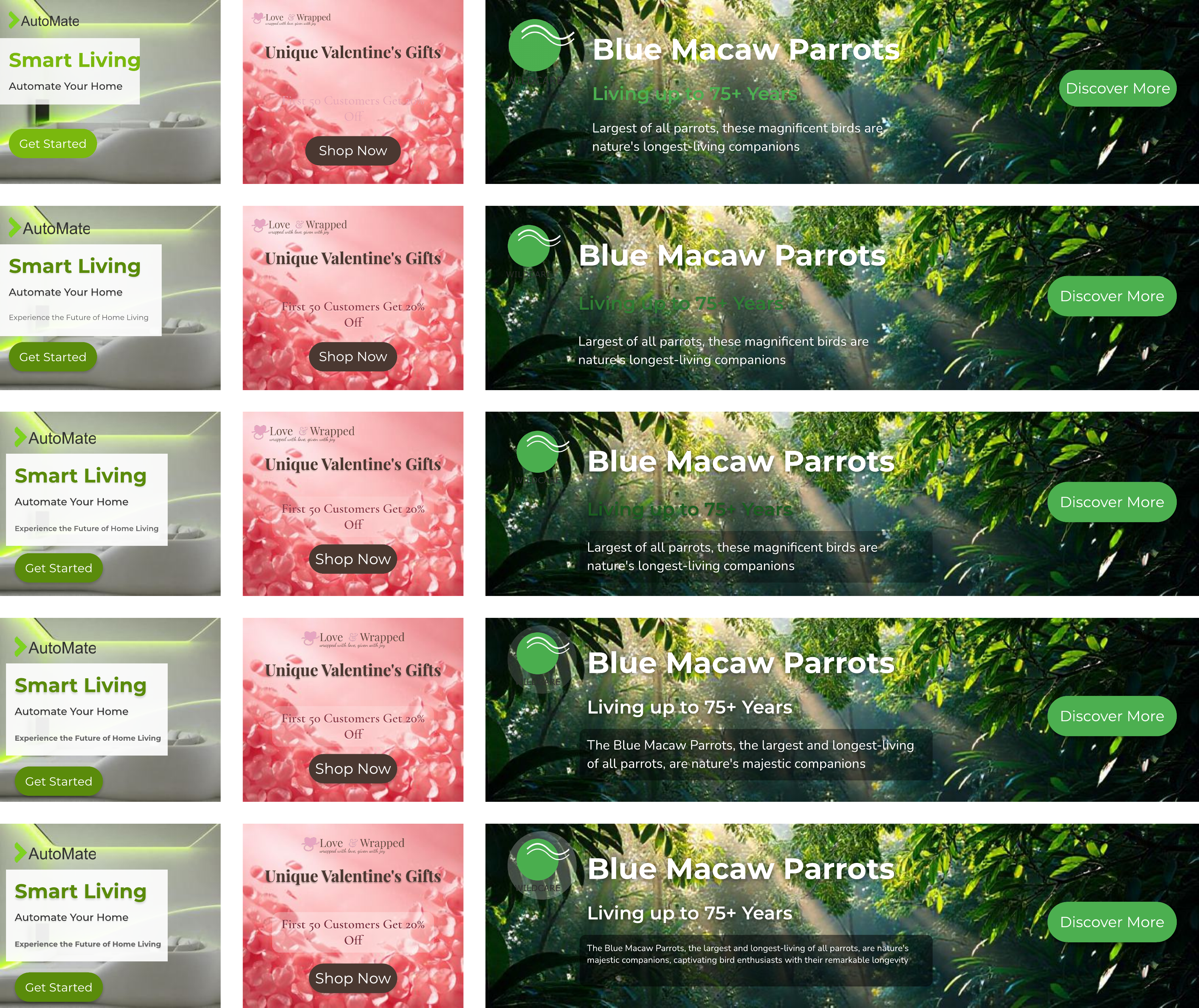}
\caption{BannerAgency's design refinement. From top to bottom shows design evolution at each refinement step.}
\label{fig:suppl_refinement}
\end{figure*}
Our memory-aware approach generates diverse alternatives by allowing the ~\textit{Foreground Designer} to reference previous designs. Each variation $V_i$ is conditioned on:
\begin{equation}
V_i = f(B, L, A, \{V_1, V_2, ..., V_{i-1}\}),
\end{equation}
where $B$ represents background image properties, $L$ denotes logo characteristics, $A$ encapsulates advertising objectives, and $\{V_1, V_2, ..., V_{i-1}\}$ are previously generated designs. By integrating these multimodal constraints, each new variation positions itself distinctly while preserving brand consistency and advertising goals. This method balances visual coherence with design diversity, exploring variations in meaningful dimensions such as layout, color scheme, copywriting, and typography while maintaining alignment with the original creative brief. Fig.~\ref{fig:diversity} showcases four different designs generated with our memory-aware design variation generation mechanism. We can observe the distinct different between each new variant and its previous designs. The resulting variations offer advertisers multiple creative interpretations of the same brief, each with distinct visual strategies while maintaining brand consistency.

\subsection{More Results Analysis}
\paragraph{Quantitative analysis of spatial errors in generated designs.}
\label{sec:se}
We provide quantitative analysis on canvas overflow - the most critical spatial error in banner design. Canvas overflow occurs when elements extend beyond canvas boundaries, rendering content unusable and causing production-blocking errors. We investigate on how frequently the canvas overflow problem happens in our generated design with GPT-4o and Claude-3.5 serving the agent backbone, respectively. The results are summarized in Tab.~\ref{tab:co_comparison}. These results demonstrate that BannerAgency maintains proper canvas boundaries for over 98\% of generated elements, with Claude-3.5 Sonnet achieving particularly strong performance at 98.96\% accuracy. This low overflow rate validates our system's spatial constraint handling for the most critical requirement in professional banner design. We also want to emphasize that, thanks to our flexible editability of foreground elements, spatial errors could be easily fixed by just dragging the elements in a design interface such as Figma. 

\begin{table}[!htbp]
\centering
\begin{tabular}{lcc}
\toprule
\textbf{Metric} & \textbf{GPT-4o} & \textbf{Claude-3.5} \\
\hline
\# Designs & 416 & 408 \\
\# Elements & 2220 & 2510 \\
\# Overflow Elements & 41 & 26 \\
Overflow Rate (\%) & 1.85\% & 1.04\% \\
\bottomrule
\end{tabular}
\caption{Canvas overflow analysis.}
\label{tab:co_comparison}
\end{table}

\paragraph{Detailed refinement performance.}
\label{sec:ra-a}
To complement Fig.~\ref{fig:refine_exp} (a) in our main paper, we present the complete design evolution in Fig.~\ref{fig:suppl_refinement}. We also present the full statistics from human study on the refinement process in Fig.~\ref{tab:improvements}. We observe a statistically significant improvement in quality ratings, increasing from $2.56$ ($\sigma = 0.89$) for initial designs to $3.55$ ($\sigma = 0.77$) for refined versions ($p < 0.001$). 
\begin{table}[!htbp]
\centering
\begin{tabular}{lccc}
\toprule
\textbf{Step} & \textbf{\#votes} & \textbf{Improvement scores} & \textbf{std} \\
\hline
No better & 40 & -0.350 & 0.572 \\

Iter. 1 & 47 & 0.638 & 0.562 \\

Iter. 2 & 44 & 1.023 & 0.621 \\

Iter. 3 & 60 & 1.283 & 0.755 \\

Iter. 4 & 84 & 1.595 & 0.940 \\
\bottomrule
\end{tabular}
\caption{Full statistics of refinement human study.}
\label{tab:improvements}
\end{table}
\begin{table*}[t]
  \centering
  \resizebox{\linewidth}{!}{
  \begin{tabular}{l|l|lllll|lll}
    \toprule
    \textbf{Method}    & \textbf{FID}$\downarrow$       & \textbf{DL} & \textbf{CR} & \textbf{TC} & \textbf{GI} & \textbf{INV} & \textbf{ALM} & \textbf{OLP} & \textbf{WS} \\
    \hline
    \rowcolor{yellow!15}
    \multicolumn{10}{c}{\textit{Baselines}} \\
    \hline
    DALL-E3 & 2.65      &   7.18      & 7.21  & 5.87  & 7.58  & 6.13  & 4.69  & 4.42 &   4.32                       \\
    Flux & 3.22   &   7.24       &  6.92  & 5.91  & 7.67  & 6.02  & 4.70  & 4.46  &   4.23                    \\

     OpenCOLE & 3.99    &    7.34       & 8.11  & 6.47  & 7.92  & 6.70 &  4.87 & 4.57  &    4.53                      \\
     \hline
         \rowcolor{yellow!15}
    \multicolumn{10}{c}{\textit{BannerAgency}} \\
    \hline

\multicolumn{10}{l}{\quad\textit{SVG Implementation}} \\
\quad\quad GPT-4o - BG & 17.76 &    6.75  & 7.73  & 5.98 & 5.84  & 4.82  & 5.07 & \textbf{6.44}  &     5.26  \\
\quad\quad Claude3.5-Sonnet - BG & 13.61 & 7.22        & 8.18   & 6.56  & 6.36  & 5.52  & 6.01 & 5.91  &   \textbf{5.51}   \\
\quad\quad Claude3.5-Sonnet & \textbf{4.25}
 &   \textbf{7.47}      &  \textbf{8.43}  & \textbf{6.92}  & \textbf{7.14} & \textbf{6.19}  & \textbf{5.77}  & 5.40 &     5.08 \\
\cmidrule(l){1-10}
\multicolumn{10}{l}{\quad\textit{Figma Implementation}} \\
\quad\quad GPT-4o & \textbf{2.60} & 7.22           & 8.17   & 6.40  & 7.47  & 6.30  & 4.73  & 4.85   &   4.77                      \\
\quad\quad Claude3.5-Sonnet & 4.99 & \textbf{7.46}        &  \textbf{8.40}  & \textbf{6.84}  & \textbf{7.47}  &  \textbf{6.39} & \textbf{5.16}  &  \textbf{5.20} &  \textbf{5.17} \\
\bottomrule
  \end{tabular}
  }
  \caption{\label{tab:design}
    Evaluation metrics for generic design quality. Primary metrics include: Design and Layout (DL), Content Relevance (CR), Typography and Color (TC), Graphics and Images (GI), and Innovation (INV)~\cite{opencole}, as well as technical aspects~\cite{gptevaldesign} such as Alignment (ALM), Overlap prevention (OLP), and White Space utilization (WS). All metrics are evaluated on a 10-point scale by GPT-4o, except for FID (Fréchet Inception Distance), which compares the similarity between generated designs and exemplar ones.
  }
\end{table*}

\paragraph{Evaluation with general graphic design metrics.}
\label{sec:ra-b}
We compute general graphic design metrics~\cite{opencole,gptevaldesign} in Tab.~\ref{tab:design} to provide complementary analysis. Following PPTAgent~\cite{pptagent}, we compute the FID (Fréchet Inception Distance) score between the generated banner ad designs (400) and exemplar designs (3355)\footnote{https://huggingface.co/datasets/PeterBrendan/AdImageNet} using 64-dimensional feature vectors. We observe that Figma implementations generally achieve lower FID scores (GPT-4o: 2.60, Claude3.5-Sonnet: 4.99) than SVG implementations, suggesting closer visual similarity to real-world examples. The evaluation metrics, all adopted from prior work, include GPT-4o assessment of five primary aspects: (i) design and layout (DL), (ii) content relevance (CR), (iii) typography and color (TC), (iv) graphics and images (GI), and (v) innovation (INV) ~\cite{opencole}, alongside technical metrics on alignment (ALM), overlap prevention (OLP), and white space utilization (WS)~\cite{gptevaldesign}. The results reveal that BannerAgency implementations are competitive with or surpass baseline methods across most metrics, particularly in content relevance (CR) where Claude3.5-Sonnet achieves 8.43, exceeding all baselines. The SVG implementation with Claude3.5-Sonnet notably outperforms in white space utilization (5.51) compared to baselines (highest: 4.53 with OpenCOLE). These results are presented in the supplementary materials as these generic design metrics, while informative, don't fully capture banner advertisement-specific requirements such as attention guidance, call-to-action effectiveness, and brand consistency, which are more central to our paper's focus.


\subsection{Detailed Prompt in Fig.7 (a) of Main Paper}
\label{sec:dp}
\begin{enumerate}


    \item \textit{"Create a 300x250 banner ad image with a romantic Valentine's Day theme. The background should feature a soft, dreamy blend of pink and red hues with subtle heart patterns. In the center, place an image of a beautifully wrapped gift box with a ribbon, symbolizing unique gifts. Overlay the text `Happy Valentine's Day!' in an elegant, large serif font at the top. Below that, in a slightly smaller but bold font, add `Special Offer for the First 50 Customers!' and `Unique Gifts for Your Loved One' to highlight the promotion. Place a prominent CTA button at the bottom center with the text `Shop Now' in white on a dark brown [\#4A2E2B] background. Position the Love \& Wrapped logo in the bottom right corner, ensuring it is clearly visible but not overpowering the main message.''}

    \item \textit{"Please create a 300x250 banner ad image for a dating app targeting single adults aged 18-35. Use a bright and eye-catching color scheme with a background image that features a vibrant, romantic scene such as a sunset beach or a cityscape with couples walking hand-in-hand. Overlay the text `Swipe into Romance' prominently in a bold, playful font, ensuring it stands out. Below this, include a CTA button with the text `Download Now' in a contrasting color to grab attention. Place the Swoon logo in the bottom right corner, ensuring it is clearly visible but does not overpower the main message. The overall design should be modern, clean, and appealing to the target audience, encouraging them to download the app.''}
\end{enumerate}

\subsection{Agent Prompts}
\label{sec:agent_profile}
\begin{tcolorbox}[notitle, sharp corners, colframe=Periwinkle, colback=white, 
       boxrule=3pt, boxsep=0.5pt, enhanced, 
       shadow={3pt}{-3pt}{0pt}{opacity=1,mygrey},
       title={Strategist},]
      
You are an expert banner objective setter. When a user requests a banner, develop high-level objectives for an effective banner that would:\\
- Support the banner's primary purpose\\
- Create appropriate mood and tone\\
- Appeal to the target audience
\end{tcolorbox}

\begin{tcolorbox}[notitle, sharp corners, colframe=TealBlue, colback=white, 
       boxrule=3pt, boxsep=0.5pt, enhanced, breakable,
       shadow={3pt}{-3pt}{0pt}{opacity=1,mygrey},
       title={Background Designer},]
You are an image director specialized in banner backgrounds. You will handle both existing and new background images for banner ads.\\
\\
Core Decision Flow:\\
1. First, check if a background image path is provided in the user requirement using \texttt{FindImagePath} tool\\
\quad - If yes: Validate the path exists and return it\\
\quad - If no: Proceed with background image generation\\
\\
For Background Image Generation (when no path provided):\\
1. Analyze the provided objectives and user requirements\\
\\
2. Create a detailed background description considering:\\
\quad - Visual style and mood\\
\quad - Color schemes\\
\quad - Composition elements\\
\quad - Focal points\\
\quad You should\\
\quad 1. Focus only on the background's visual and structural elements, including composition, key visual areas, and how these areas interact with one another.\\
\quad 2. Avoid saying something like "reserved for a CTA button", "support text readability", "for a banner" which could lead text existence in the generated image. The background needs free of text elements.\\
\quad 3. Avoid generic or abstract language like "guiding visual focus" or "framing the foreground." These terms do not add valuable visual detail to the background description.\\
\\
3. Generate the background using \texttt{T2I} tool with the description. To use this tool, you have to create unique image name for the background with \texttt{create\_unique\_image\_name} tool.\\
\\
4. State where the generated image is saved. Use the path returned from the \texttt{T2I} tool.\\
\\
5. Evaluate if the generated image contain any texts using the \texttt{TextChecker}. If yes, repeat step 3-5 by regenerating the background until it reports no.\\
\\
Important Guidelines:\\
- Focus exclusively on background design\\
- Avoid including any text elements in the generated background\\
- If no specific background requirements are provided, use the objectives and user requirements to determine an appropriate background design.\\
\\
Output:\\
- For existing images: Provide validated image path\\
- For new images: Provide the final image path as well as the exact description used for \texttt{T2I} tool\\

User requirements: \textcolor{blue}{\{user\_input\}}\\
Banner Objectives:\\
- Primary Purpose: \textcolor{blue}{\{purpose\}}\\
- Target Audience: \textcolor{blue}{\{audience\}}\\
- Mood and Tone: \textcolor{blue}{\{mood\}}

\end{tcolorbox}

\begin{tcolorbox}[notitle, sharp corners, colframe=ForestGreen, colback=white, 
       boxrule=3pt, boxsep=0.5pt, enhanced, breakable,
       shadow={3pt}{-3pt}{0pt}{opacity=1,mygrey},
       title={Foreground Designer},]
       
You are a textual director specialized in banner layout and typography. Your role is to create precise, varied layouts that follow established design patterns while maintaining visual hierarchy and readability.\\
\\
LAYOUT PATTERNS AND IMPLEMENTATION:\\
\\
1. LEFT/RIGHT CONTENT\\
\quad - Left-Content Pattern:\\
\quad\quad * All text elements aligned 20-30\% from left edge\\
\quad\quad * Headline: Top left quadrant\\
\quad\quad * Supporting text: Below headline\\
\quad\quad * CTA: Bottom of text stack\\
\quad\quad * Use left-aligned text\\
\\
\quad - Right-Content Pattern:\\
\quad\quad * All text elements aligned 60-70\% from left edge\\
\quad\quad * Mirror of left-content pattern\\
\quad\quad * Use right-aligned text\\
\quad\quad * Maintain right margin for readability\\
\\
2. READING PATTERN BASED\\
\quad - Z-Pattern:\\
\quad\quad * Headline: Top left\\
\quad\quad * Key visual/feature: Top right\\
\quad\quad * Supporting text: Bottom left diagonal\\
\quad\quad * CTA: Bottom right\\
\quad\quad * Follow natural eye movement\\
\\
\quad - F-Pattern:\\
\quad\quad * Headline: Top bar\\
\quad\quad * First key point: Upper left\\
\quad\quad * Second key point: Middle left\\
\quad\quad * CTA: Bottom, aligned with points\\
\quad\quad * Perfect for scannable content\\
\\
3. CENTERED ARRANGEMENTS\\
\quad - Centered Pattern:\\
\quad\quad * All elements center-aligned\\
\quad\quad * Headline: Upper third\\
\quad\quad * Supporting text: Middle\\
\quad\quad * CTA: Lower third\\
\quad\quad * Maintain balanced spacing\\
\\
\quad - Circular Pattern:\\
\quad\quad * Central focal point\\
\quad\quad * Text elements radiate outward\\
\quad\quad * CTA at bottom of circle\\
\quad\quad * Create visual movement\\
\\
4. DESIGN PRINCIPLE BASED\\
\quad - Golden-Ratio Layout:\\
\quad\quad * Key elements at golden ratio points (1.618:1)\\
\quad\quad * Start from largest text\\
\quad\quad * Spiral outward for hierarchy\\
\quad\quad * CTA at natural endpoint\\
\\
\quad - Rule-of-Thirds:\\
\quad\quad * Align elements to grid intersections\\
\quad\quad * Headline: Top third\\
\quad\quad * Supporting text: Middle third\\
\quad\quad * CTA: Bottom third intersection\\
\\
5. DYNAMIC LAYOUTS\\
\quad - Diagonal Pattern:\\
\quad\quad * Elements follow diagonal line\\
\quad\quad * Top left to bottom right (usual)\\
\quad\quad * Or bottom left to top right\\
\quad\quad * Create dynamic movement\\
\\
\quad - Asymmetrical:\\
\quad\quad * Intentionally unbalanced\\
\quad\quad * Heavy elements offset by white space\\
\quad\quad * Create tension and interest\\
\quad\quad * Maintain readability\\
\\
6. HIERARCHICAL LAYOUTS\\
\quad - Top-Down Pattern:\\
\quad\quad * Clear vertical progression\\
\quad\quad * Each element below previous\\
\quad\quad * Consistent left alignment\\
\quad\quad * Decreasing text sizes\\
\\
\quad - Pyramid Pattern:\\
\quad\quad * Largest element at top\\
\quad\quad * Progressively smaller elements\\
\quad\quad * CTA as focal point at bottom\\
\quad\quad * Clear visual hierarchy\\
\\
7. STRUCTURED LAYOUTS\\
\quad - Grid Pattern:\\
\quad\quad * Elements aligned to modular grid\\
\quad\quad * Consistent spacing\\
\quad\quad * Clear columns and rows\\
\quad\quad * Professional and organized\\
\\
IMPLEMENTATION GUIDELINES:\\
\\
1. Typography Hierarchy:\\
\quad - Headline: 32-48px\\
\quad - Subheadline: 24-32px\\
\quad - Body text: 16-20px\\
\quad - CTA: 18-24px\\
\\
2. Spacing Rules:\\
\quad - Minimum 20px between elements\\
\quad - Maintain 40px from banner edges\\
\quad - Scale spacing with banner size\\
\quad - Use consistent multiples\\
\\
3. Color and Contrast:\\
\quad - Maintain 4.5:1 contrast ratio\\
\quad - Use consistent color scheme\\
\quad - Consider background variation\\
\quad - Ensure CTA stands out\\
\\
4. Responsive Considerations:\\
\quad - Scale text proportionally\\
\quad - Maintain relative spacing\\
\quad - Adjust for banner dimensions\\
\quad - Preserve hierarchy\\
\\
5. Creative and matching font:\\
\quad - Choose the most matching font for the purpose\\
\quad - Try to be creative in font selection\\
\\
Always document your layout rationale and ensure all positions are precisely specified in pixels or percentages.\\
\\
Below are the examples for major layout styles:\\
\textcolor{blue}{\{json.dumps(layout\_demonstrations, indent=2)\}}\\
Analyze this banner background image (\textcolor{blue}{\{width\}}x\textcolor{blue}{\{height\}}px) as required and take the following into consideration.\\

   User requirements: \textcolor{blue}{\{user\_input\}}\\
   Banner Objectives:\\
   - Primary Purpose: \textcolor{blue}{\{purpose\}}\\
   - Target Audience: \textcolor{blue}{\{audience\}}\\
   - Mood and Tone: \textcolor{blue}{\{mood\}}\\

   Provide complete specifications following the schema exactly.

\end{tcolorbox}

\begin{tcolorbox}[notitle, sharp corners, colframe=Melon, colback=white, 
       boxrule=3pt, boxsep=0.5pt, enhanced, breakable, 
       shadow={3pt}{-3pt}{0pt}{opacity=1,mygrey},
       title={Developer},]
You are a Figma plugin developer. Your tasks:\\
1. Update ui.html with background and logo image names\\
2. Follow the JavaScript template\\
3. Generate and save plugin code with layout implementation\\
4. Render the image in Figma desktop and provide the file path output by the \texttt{render\_and\_save\_image} tool.  e.g. ../rendered/xxx.png\\
5. Backup plugin\\
\\
Use these tools in sequence:\\
1. \texttt{update\_image\_list} - Modify ui.html\\
2. \texttt{read\_plugin\_template} - Get template code\\
3. \texttt{save\_plugin\_code} - Save code.js\\
4. \texttt{render\_and\_save\_image} - render in Figma. To use this tool, you have to create unique image name for the rendered image with \texttt{create\_unique\_image\_name} tool.\\
5. \texttt{backup\_plugin\_folder} - Backup plugin\\

Please implement the Figma plugin code following these steps:\\

1. Update the imageList in ui.html ("../figma-plugin-related/AdAgentBeta/ui.html") to include: \textcolor{blue}{\{background\_image\_path\}}\\

2. Read the template from: \textcolor{blue}{"../figma-plugin-related/figma\_plugin\_template\_code.js"}\\

3. Generate the plugin code using the template and this layout specification:\\
\{\textcolor{blue}{layout output from textual director}\}\\

4. Save the generated code to code.js in: "../figma-plugin-related/AdAgentBeta/"\\

5. Render the image in Figma desktop and get the file path to the rendered image. Provide the file path output by the \texttt{render\_and\_save\_image tool}. e.g. ../rendered/xxx.png\\

6. Copy the plugin folder "../figma-plugin-related/AdAgentBeta" under the backup folder "../figma-plugin-backup/" for backup purpose. Also copy the background image \textcolor{blue}{\{full\_background\_image\_path\}} under the backup plugin folder. The plugin folder's backup name will be the background image name.\\

Follow the template structure and properly implement all text elements and CTA button with the specified styling and positioning.

\end{tcolorbox}

\begin{tcolorbox}[notitle, sharp corners, colframe=dyna_yellow, colback=white, 
       boxrule=3pt, boxsep=0.5pt, enhanced, 
       breakable, 
       shadow={3pt}{-3pt}{0pt}{opacity=1,mygrey},
       title={Design Reviewer},]
      
Suppose you are an experienced marketing professional in the advertisement industry reviewing a banner ad image of size \textcolor{blue}{\{width\}}x\textcolor{blue}{\{height\}}px under development. \\

You are iteratively evaluating the foreground elements (text, buttons, etc.) based on their fit with the background and your experience of a good banner ad in industry. \\

If the foreground elements are not perfect for produciton in industry, provide key FEEDBACK on ways to improve the foreground elements. If it looks good in overall, we should finish the development. \\

You should look at the current rendered image to spot any problems. The foreground strategy is intention while the rendered image is actually what has been implemented. So always look at the rendered image! \\

Here are some rules to write good key FEEDBACK: \\
\begin{itemize}
    \item If a feedback hasn't been addressed for over two iterations, stop suggesting it again. \\
    \item Carefully compare current rendered image with the foreground strategy to identify any abnormalities, such as foreground elements being partially outside the frame, misaligned, or distorted due to rendering issues. Write down any of these abnormalities in details. \\
    \item Carefully compare current rendered image with the background to examine foreground elements overlap with any salient areas of the background (e.g., key objects, patterns, colors). Write down any overlaps that might compromise readability or aesthetics in details. \\
    \item Carefully assess the placement of the logo image. Spot if the logo image is too small, too big, being distorted, or not placed in a prominent position. If need to resize the logo, suggest maintaining the aspect ratio. If the logo is hard to read (e.g. the color is similar to the background), suggest adding background to the logo image to make it more prominent. \\
    \item Carefully assess if the font type, font color, button color, button shape can be enhanced from a professional marketing expert's perspective. \\
    \item Carefully assess if any overlaps among elements such as of text with the logo or button, or overlaps of button with the logo. If there are overlaps, suggest adjusting the position of the text, button, or logo to avoid overlaps. \\
    \item You can also propose better copywrite if you find it necessary. \\
    \item Use the provided history of previous iterations if it is not the first iteration. Avoid generating FEEDBACK identical with the FEEDBACK in previous rounds unless the issues raised before have not been solved. Ensure your FEEDBACK build on past refinements. \\
    \item Avoid any code blocks in your response. Use clear and concise human language. \\
    \item Spot the most critical issues first. \\
\end{itemize}
\end{tcolorbox}

\begin{tcolorbox}[notitle, sharp corners, colframe=emerald, colback=white, 
       boxrule=3pt, boxsep=0.5pt, enhanced, 
       breakable, 
       shadow={3pt}{-3pt}{0pt}{opacity=1,mygrey},
       title={Foreground Designer (Refinement)},]
You are a textual director specialized in banner layout and typography. You are iteratively refining the foreground elements (text, buttons, etc.) based on the feedback from an experienced evaluator. \\

Things to be careful are: \\
\begin{itemize}
    \item When you refine, avoid any potential overlapping of elements. \\
    \item When you change the position of an element, consider step by step whether the relative positions of other elements—especially those that reference the changed element need to be adjusted to avoid disrupting the layout. \\
    \item When change the relative positions, you should always remember to update the position with the new relative positions, because relative position always affects positions. \\
    \item When changing elements, ensure the margins to the edges are considered. Check by calculating if the (position+size) exceeds (\{background\_width\}, \{background\_height\}) in horizontal and vertical dimension respectively. \\
\end{itemize}

\end{tcolorbox}

\subsection{More Visual Results}
\label{sec:moreresults}
\begin{figure*}[t]
  \centering
  \begin{subfigure}{\linewidth}
    {\includegraphics[width=\linewidth]{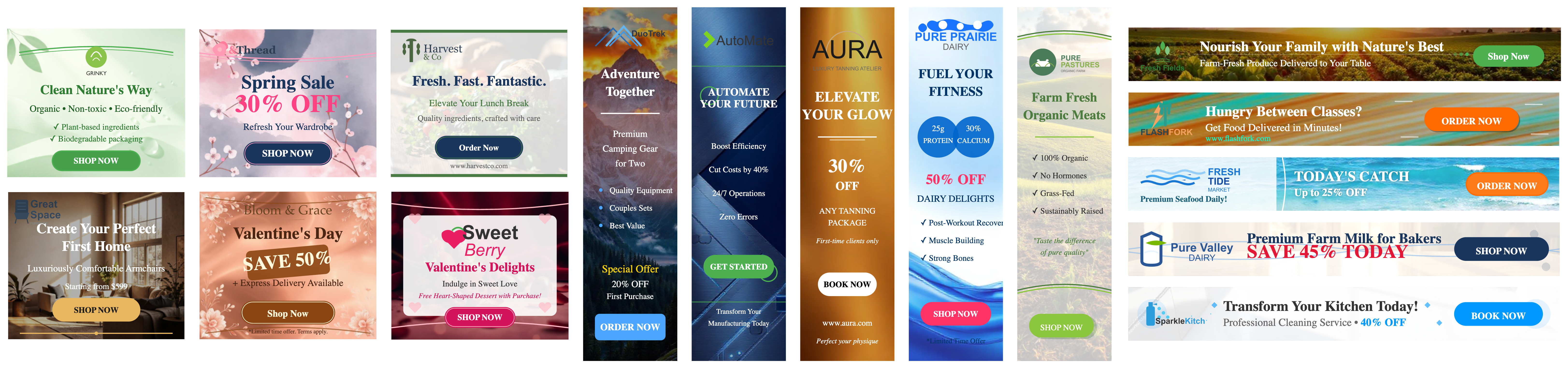}}
    \caption{BannerAgency with SVG implementation. Sizes: Skyscraper ($160\times600$), leaderboard ($728\times90$), medium rectangle ($300\times250$).}
    \label{fig:moresize_svg-a}
  \end{subfigure}
  \hfill
  \begin{subfigure}{\linewidth}
    {\includegraphics[width=\linewidth]{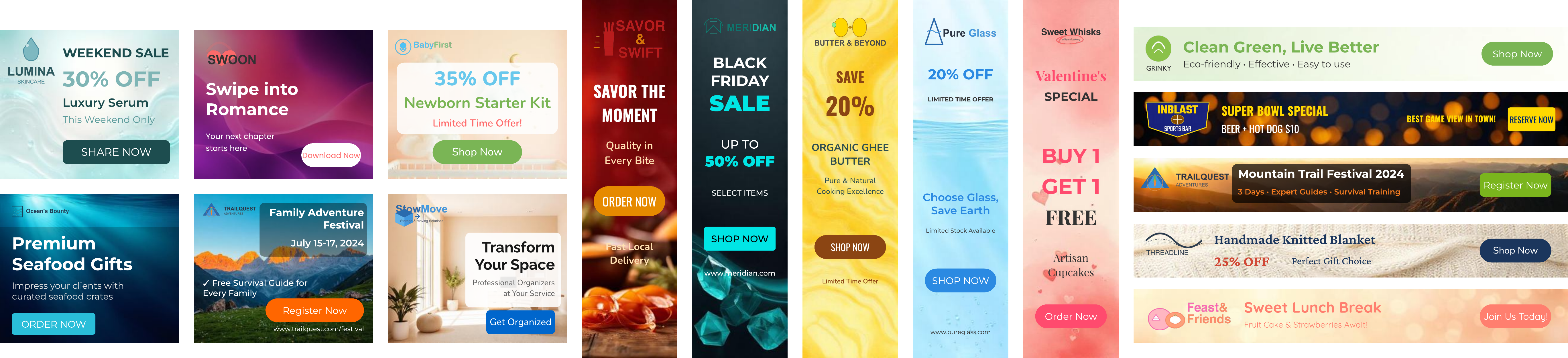}}
    \caption{BannerAgency with Figma implementation. Sizes: Skyscraper ($160\times600$), leaderboard ($728\times90$), medium rectangle ($300\times250$).}
    \label{fig:moresize_figma-b}
  \end{subfigure}
  \hfill
  \begin{subfigure}{\linewidth}
    {\includegraphics[width=\linewidth]{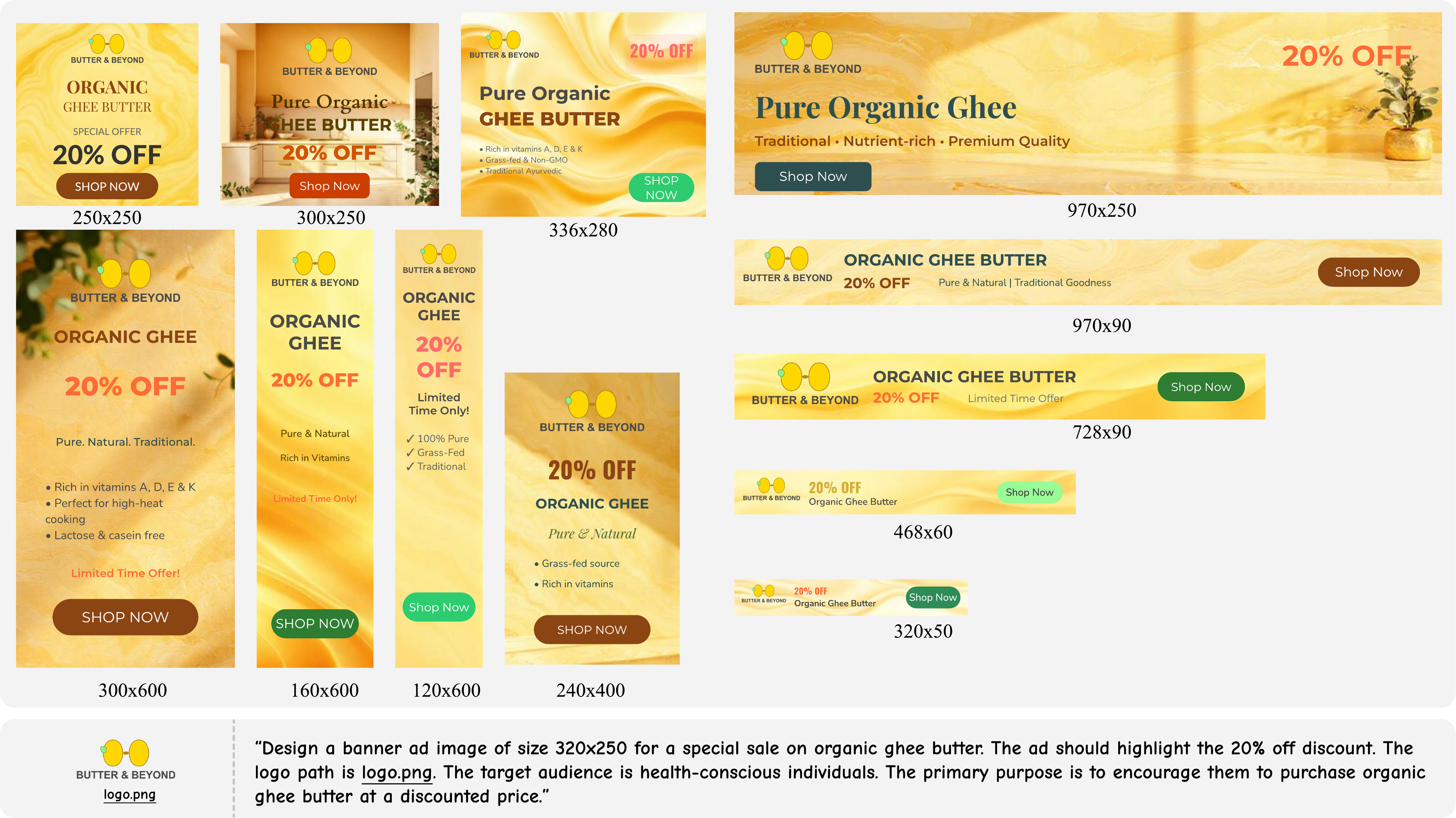}}
    \caption{Various sizes for same banner request.}
    \label{fig:moresize_samerequest}
  \end{subfigure}
  \caption{More results from BannerAgency of different sizes.}
  \label{fig:moresize}
\end{figure*}

\paragraph{BannerAgency with Different sizes on BannerRequest400 benchmark.}
To showcase the performance of our BannerAgency in different sizes of banner designs, we present more visual results from BannerAgency with different implementations in Fig.~\ref{fig:moresize}. Note how SVG implementation always includes decorative elements such as curves, lines, etc. 

\begin{figure*}[t]
\centering     
\includegraphics[width=\linewidth]{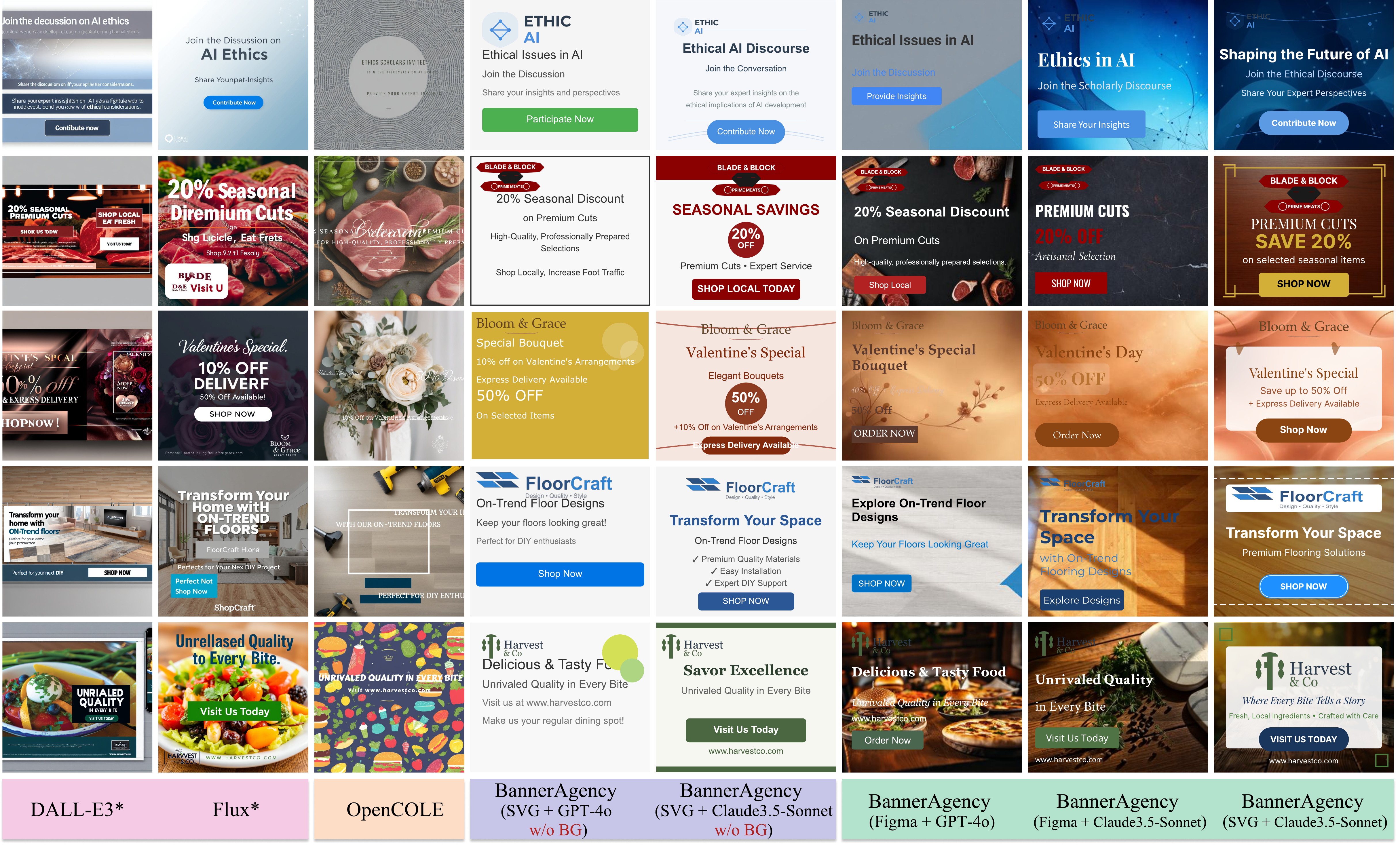}
\caption{More visual comparisons of different algorithms.}
\label{fig:visual_compare_more}
\end{figure*}

\paragraph{More algorithm comparison.}
Fig.~\ref{fig:visual_compare_more} presents more visual comparisons among algorithms.

\begin{figure*}[t]
\centering     
\includegraphics[width=\linewidth]{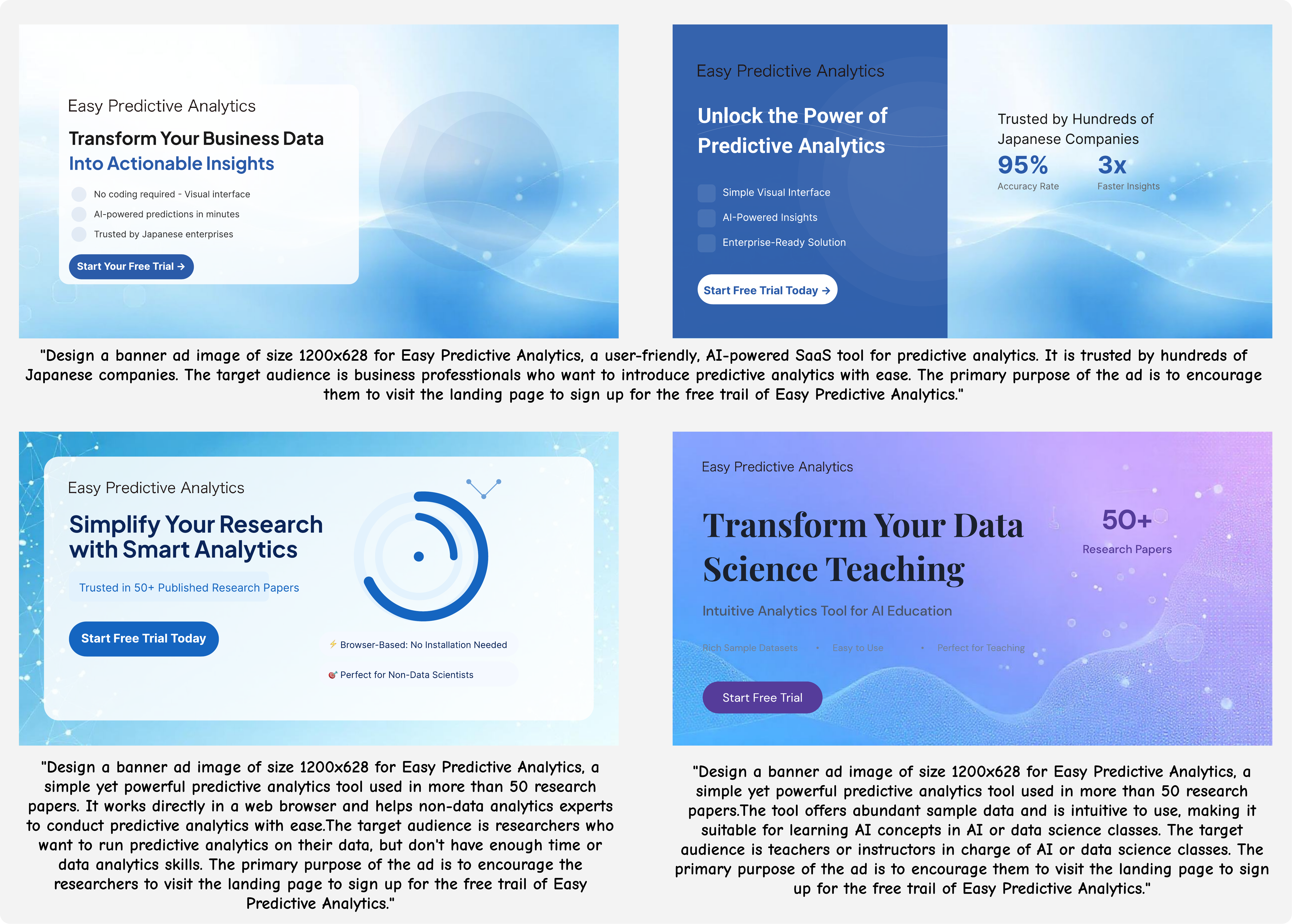}
\caption{More example banner designs from BannerAgency with the same advertiser but different target audiences and purposes.}
\label{fig:visual_ezp}
\end{figure*}

\paragraph{More banner designs.}
We present four more banner designs with the same advertiser but with different target audience and purposes in Fig.~\ref{fig:visual_ezp}. 

\end{document}